%% file: main.tex
\documentclass[10pt,twocolumn,letterpaper]{article}

\usepackage[pagenumbers]{cvpr} 
\usepackage{graphicx}
\input{preamble}

\definecolor{cvprblue}{rgb}{0.21,0.49,0.74}
\usepackage[pagebackref,breaklinks,colorlinks,citecolor=cvprblue]{hyperref}

\def\paperID{10619} 
\def\confName{CVPR}
\def\confYear{2024}

\title{MuseChat: A Conversational Music Recommendation System for Videos}

\author{
Zhikang Dong \thanks{Corresponding author.} \thanks{Equally contributed. Work partially done during the internship at TikTok.} \\
Stony Brook University\\
{\tt \small zhikang.dong.1@stonybrook.edu}
\and
Xiulong Liu \footnotemark[2] \\
University of Washington\\
{\tt \small xl1995@uw.edu}
\and
Bin Chen \\
Bytedance.com\\
{\tt \small chen.bin@bytedance.com}
\and
Pawe\l \ Polak \\
Stony Brook University\\
{\tt \small pawel.polak@stonybrook.edu}
\and
Peng Zhang \\
Bytedance.com\\
{\tt \small zhang.peng@bytedance.com}
}

\begin{document}
\maketitle
\input{sec/abstract}    
\input{sec/intro}
\input{sec/related_work}

\input{sec/dataset_1}
\input{sec/method}
\input{sec/results}
\input{sec/conclusions}
{
    \small
    \bibliographystyle{ieeenat_fullname}
    \bibliography{main}
}

\input{sec/X_suppl}

\end{document}

%% file: preamble.tex
\usepackage[dvipsnames]{xcolor}


%% file: sec/abstract.tex
\begin{abstract}
Music recommendation for videos attracts growing interest in multi-modal research. However, existing systems focus primarily on content compatibility, often ignoring the users' preferences. Their inability to interact with users for further refinements or to provide explanations leads to a less satisfying experience.
We address these issues with MuseChat, a first-of-its-kind dialogue-based recommendation system that personalizes music suggestions for videos. Our system consists of two key functionalities with associated modules: recommendation and reasoning. The recommendation module takes a video along with optional information including previous suggested music and user's preference as inputs and retrieves an appropriate music matching the context. The reasoning module, equipped with the power of Large Language Model (Vicuna-7B) and extended to multi-modal inputs, is able to provide reasonable explanation for the recommended music. To evaluate the effectiveness of MuseChat, we build a large-scale dataset, conversational music recommendation for videos, that simulates a two-turn interaction between a user and a recommender based on accurate music track information. Experiment results show that MuseChat achieves significant improvements over existing video-based music retrieval methods as well as offers strong interpretability and interactability.

\end{abstract}

%% file: sec/intro.tex
\section{Introduction}
\label{sec:intro}

Music is an essential component in videos, enhancing both the viewer's experience and their understanding of the content. While existing music recommendation systems are proficient at selecting tracks that align with a video's theme—such as scary music for a horror film or upbeat tunes for a dance clip—these systems often neglect individual user preferences. For example, an `80s enthusiast may favor synth-pop over modern pop music for a nostalgia-themed video, even though both are categorized as ``pop.''

\begin{figure*}[t]
  \centering
   \includegraphics[width=0.9\linewidth]{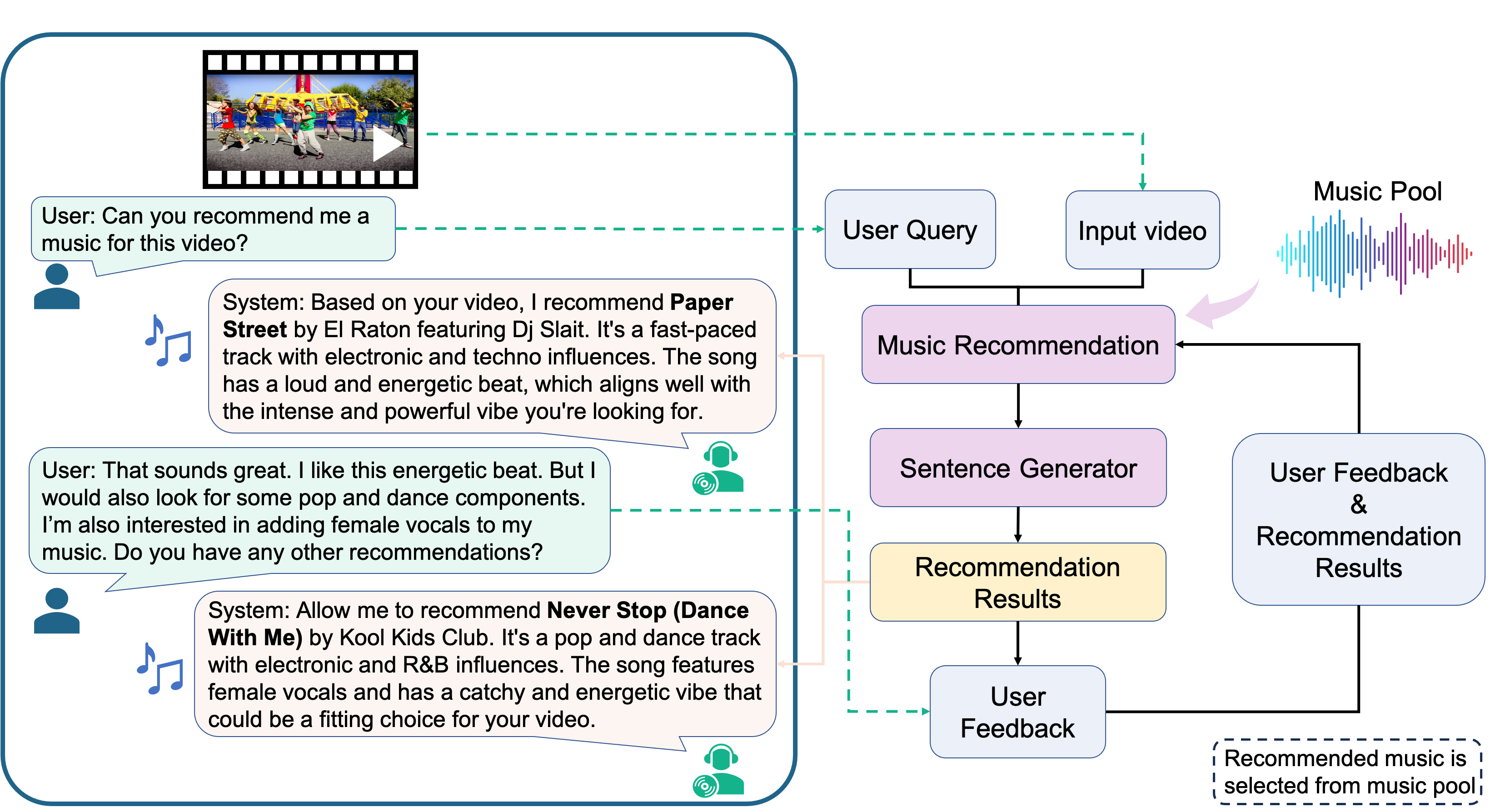}

   \caption{MuseChat features two modules: the Music Recommendation Module, which processes either video input alone or in combination with user prompts and past music suggestions, and the Sentence Generator Module, which uses these inputs to create natural language music recommendations.}
   \label{fig:overall_pipeline}
\end{figure*}

The challenge of personalized recommendation remains relevant, as many systems leverage user profiles and activity data to generate recommendations. However, we identify two key limitations: (1) the inability to consistently meet user preferences, and (2) the cold-start problem for new users without prior data. Current music recommendation systems aim to provide lists of songs based on user history, but these may not always align with user needs for specific videos. This not only affects user experience but also underscores the complexity of predicting preferences, which may deviate due to factors like user's recent trends. We propose a feedback mechanism to mitigate these limitations. This would allow the system to adjust its recommendations according to user feedback, aligning more closely with changing preferences. For new users without historical data, a cold-start scenario arises, leading to content-driven recommendation once again.

In this study, we introduce MuseChat, a comprehensive conversational music recommendation system for videos. As shown in Figure \ref{fig:overall_pipeline}, MuseChat consists of two main modules: the music recommendation module and the sentence generator module. Users begin with uploading a video and receive a music recommendation tailored to the video's content. MuseChat enables user-system interactions through dialogues in natural language. At each dialogue turn, users are empowered by the music recommendation module in MuseChat. They can refine recommendations by specifying their preferences in natural language, such as mood, genre, instruments, theme, and artist details. This process continues until they identify their desired music track. Another distinguishing feature is the system's interpretability versus the ``black box" nature of conventional music recommendation models. Our sentence generator module can not only justify the recommended music with reasons but also craft personal narratives for users based on the selected music.

Building upon this framework, our contributions are as follows:
    (1) We introduce a large-scale dataset tailored for a novel task, dialogue-driven music recommendations and reasoning within the context of videos. The data contains 98,206 quartets: a video, original music, candidate music and a two-turn conversation. This setup mimics the user's interaction with recommendation systems. It starts with uploading a video, receiving an initial music recommendation, and then accommodating a user's textual prompt to finalize the music selection;
    (2) We present a tri-modal architecture designed for music-video matching that incorporates with textual input. This model not only processes the previously recommended music and video content but also refines recommendations based on user-provided textual prompts;
    (3) We augment our model with the capability to offer clear, logical explanations for its music recommendations, achieved through the deep understanding of musical features by our LLM-based sentence generation module.

%% file: sec/related_work.tex
\section{Related Work}
\label{sec:related}

{\bf Automatic music tagging.}
Music tags efficiently summarize songs by providing descriptive keywords that cover various elements such as emotion, genre, and theme. Numerous studies dive into the domain of automatic music tagging, as evidenced by works such as \cite{choi2016automatic,lee2020disentangled,lee2017sample,pons2019musicnn,choi2019zero,won2020evaluation}. Specifically, \cite{won2021semi} employs a model that uses shallow convolutional layers to extract acoustic features, which are then processed by stacked self-attention layers in a semi-supervised setting. Similarly, \cite{zhao2022s3t} introduces S3T, a self-supervised pre-training method based on the Swin Transformer \cite{liu2021swin} architecture, further optimized by a music-specific data augmentation process.

\noindent{\bf Music description in free-form natural language.}
Describing music in free-form natural language has also gained research attention \cite{huang2022mulan,manco2022contrastive,hu2023audio,won2020evaluation}. For instance, \cite{doh2023toward} proposes a universal retrieval system to handle both tag- and sentence-level inputs. This system demonstrates adaptability across nine different music classification tasks. Moreover, \cite{manco2022song} introduces 'Song Describer,' an open-source tool designed to gather text descriptions of music tracks from users. This initiative has led to the creation of a public audio-caption dataset in the music domain.

\noindent{\bf Music recommendation for video.}
The task of music recommendation based on video attributes receives attention in recent studies \cite{pretet2021cross, suris2022time, zeng2018audiovisual, yi2021cross}. While some works focus on creating joint embeddings of music and free-form natural language \cite{huang2022mulan, manco2022contrastive}, other studies examine the relationship between video, everyday audio sounds (excluding music), and language \cite{guzhov2022audioclip, wu2022wav2clip}. \cite{mckee2023language} develops a method for guiding music recommendations with a single text description of music attributes. However, their approach does not incorporate user feedback, nor does it adapt its recommendations based on such feedback or prior recommendation results. Our work with MuseChat aims to address these limitations.

\noindent{\bf Conversational recommendation system.}
Conversational Recommender Systems (CRS) gain research attention for their ability to support task-oriented, multi-turn dialogues with users \cite{jannach2021survey,gao2021advances,deng2023unified,yang2021improving,gao2023chat}. These systems capture the user's detailed and current preferences, provide explanations for suggested items, and process user feedback on recommendations. The emergence of LLMs substantially enhances the capabilities of CRS, particularly in understanding and generating natural language. \cite{wang2023rethinking} proposes an interactive evaluation approach that balances the focus between matching ground truth and maintaining interactivity. \cite{friedman2023leveraging} utilizes LaMDA \cite{thoppilan2022lamda} to create a YouTube video recommendation system, relying on textual metadata instead of visual information.

\noindent{\bf Multi-modalities and Large language models.}
The rapid evolution of LLMs becomes a game-changer in the landscape of artificial intelligence, becoming a focal point in contemporary research. Originating from transformer architectures \cite{vaswani2017attention}, these models train on extensive corpora, containing billions of words \cite{devlin2018bert, radford2018improving}. Noteworthy models like OpenAI's GPT-3 \cite{brown2020language}, Meta's LLaMA \cite{touvron2023llama}, and Google's LaMDA \cite{thoppilan2022lamda} set benchmarks for textual generation that closely resembles human articulation. Recent advances in LLMs extend beyond text to multi-modal inputs. These models are proficient at synthesizing and interpreting information across different data types. For instance, \cite{zhu2023minigpt} introduces MiniGPT-4, which incorporates a visual encoder into a large language model. This leads to the model's ability to generate narratives inspired by images. Other notable works include \cite{zhang2023video}, which can interpret video content to generate informed textual responses, and \cite{lin2023vision}, which demonstrates how to encode answer candidates into GPT-3 prompts, enabling external knowledge integration. \cite{liu2023tackling} proposes a novel method to answer audio-visual related questions in their balanced audio-visual-text dataset. Recent developments such as fine-tuning adapters \cite{zhang2023llamaadapter, gao2023llamaadapterv2, hu2021lora, dettmers2023qlora} make it easier for smaller research groups to adapt large models for specific uses, overcoming the high computational costs.

%% file: sec/dataset_1.tex
\section{Dataset}
\label{sec:dataset}
\begin{figure*}[!h]
  \centering
   \includegraphics[width=1\linewidth]{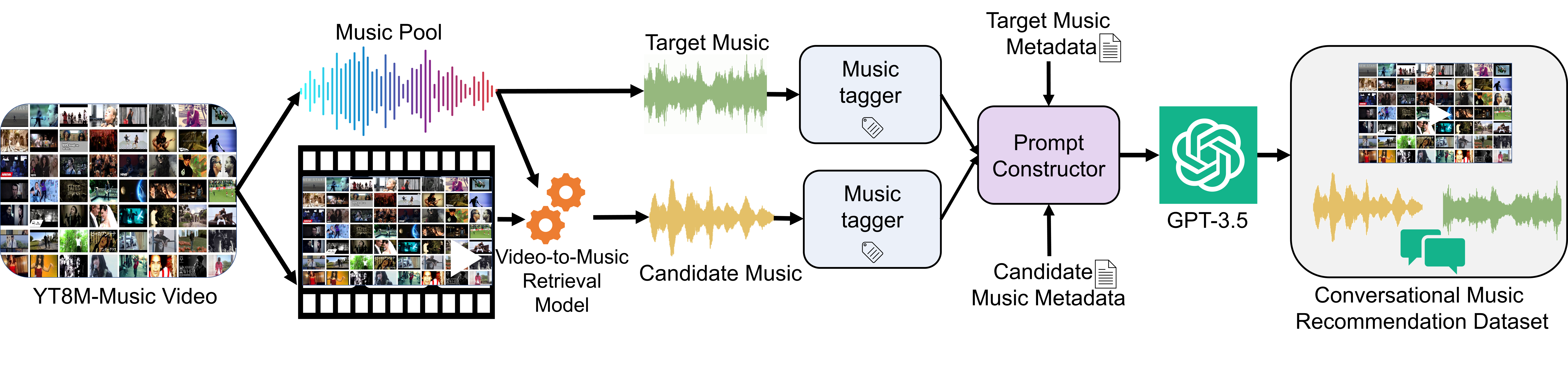}
   \caption{The generation pipeline for Conversational Music Recommendation Dataset.}
   \label{fig:data_generation}
\end{figure*}

We create the conversation part of the dataset by simulating a two-turn dialogue. In the first turn, the user provides only the video, and then a pretrained video-to-music retrieval system suggests a candidate music. In the second turn, based on this candidate music, the user provides more specific preferences in natural language as a prompt to guide the recommendation system. With the video, the candidate music, and the prompts, the system suggests the target music. While we only focus on generating two-turn dialogues here, we claim that our approach could easily extend to more dialogue turns. This is because our approach can treat subsequent dialogue turns as a ``second turn", continuously adapting recommendations based on the previous result and current user instructions.
Our approach involves the following steps, as \Cref{fig:data_generation} shows: (1) collecting a large number of video and music pairs from an existing data source; (2) using a pretrained video-to-music retrieval model to select a candidate music for each video; (3) gathering music tags and other metadata for both the original and candidate music, which are then used to construct prompts; (4) feeding these prompts into GPT-3.5 for dialogue generation. We illustrate each step separately below.

{\bf Video-music pairs collection}.
We use the YouTube8M dataset \cite{abu2016youtube} to create our conversational music recommendation dataset. YouTube8M is a large-scale labeled video dataset, encompassing millions of YouTube video IDs and relevant features for audio and visual content. It comes with thousands of labels covering a wide range of categories, such as music, sports, documentaries, and more.
We select videos labeled as ``music video" and remove unavailable videos, resulting in a collection of 98,206 music videos. From each music video, we extract a 120-second clip from the center. We choose this 120-second segment for three reasons: (1) It often captures the core information of the video, serving as a representative sample for music recommendation; (2) It minimizes noise from musical elements like intros and outros that could negatively impact the recommendation process; (3) It reduces computational costs.

{\bf Preparing candidate music}. We implement a video-to-music retrieval model, named as Music-Video Pretrained (MVP) model, to retrieve an appropriate candidate music from the music pool to pair with the video. The MVP is a two-tower model, sharing a similar architecture with the models described in \cite{huang2022mulan, suris2022time}. It takes raw video and music clips as inputs, utilizing the pretrained CLIP Image encoder \cite{radford2021learning} for video feature extraction and the pretrained Audio Spectrogram Transformer (AST) \cite{gong2021ast} for music feature extraction.\footnote{The pretrained weights used are clip-vit-large-patch14 for the CLIP Image encoder and MIT/ast-finetuned-audioset-10-10-0.4593 for the AST.} This model is trained on our proprietary dataset consisting of 3 million high quality music-video pairs. More details for MVP are included in supplementary materials. We randomly divide the 98,206 music tracks from the music video clips into non-overlapping pools, with each containing 2,000 tracks. We employ the MVP model to select the candidate music from the same pool in which target track resides but excludes the target track. We empirically find that setting the pool size to 2000 could ensure the quality and diversity of candidate music as well as stay different from the target track. The MVP model computes the similarity between the input video and each track in the pool, ranking the tracks by descending similarity to determine the candidate music.

{\bf Constructing prompts}. Our objective in this step is to construct prompts that effectively incorporate both the target and the candidate music tracks. Since each video from YouTube8M ``music video'' category has a paired music, we use it as the target music track in our simulated dialogue, towards which the user prompt would ``steer'' the recommendation from the candidate music.
We employ the music tagging method from \cite{pons2019musicnn} to assign top 5 tags to each target music track. We leverage two distinct tagging systems for this purpose: one from the MagnaTagATune (MTT) dataset \cite{güçlü2016brains} and another from the Million Song dataset (MSD) \cite{Bertin-Mahieux2011}. Each system has a vocabulary of 50 tags, and the use of both systems increases the robustness of the tagging. 
The details and statistics of music tags in our dataset are provided in the supplementary material.
Alongside music tags, we also collect metadata for music videos, which includes video title, video description, track name, artists, albums, and etc. These metadata are downloaded from the YouTube website for corresponding videos. Although each music video has a title and description, supplementary details such as official artist names, album specifics, and release dates are available for only around 30,000 tracks. We \textit{manually} create various prompt templates using music tags and metadata from both original and candidate music tracks to diversify the synthetic conversations.

{\bf Dialogue turn generation}.
In the final stage, we generate dialogues that bridge the gap between the target music and candidate music. Utilizing GPT-3.5, we process the prompts constructed in the previous step.
We observe that GPT-3.5 can effectively leverage its extensive knowledge base to enrich the conversations, even when the music metadata is unavailable. This integration of our carefully crafted prompts and GPT-3.5's generative capabilities ensures that our training data is both high-quality and varied, successfully imitating human interactions in the music recommendation context. The example prompt is available in the supplementary materials. 

%% file: sec/method.tex
\section{Approach}
\label{sec:method}
We propose a novel approach to address the conversational music recommendation task on this dataset, establishing a new baseline performance. As shown in \Cref{fig:overall_pipeline}, there are two main modules involved in the system: music recommendation and sentence generator. We illustrate them below.

\subsection{Music Recommendation}
The goal of the music recommendation module is to select the most relevant music from the music pool using video, candidate music, and user text prompt.
Each training sample is defined as a quartet $(v, m_c, m_t, t_3)$, where $v$ is the video, $m_c$ denotes the candidate music track, $m_t$ is the target music track and $t_3$ is the text of user prompt showing preferences.
\begin{figure*}[t]
  \centering
   \includegraphics[width=1.0\linewidth]{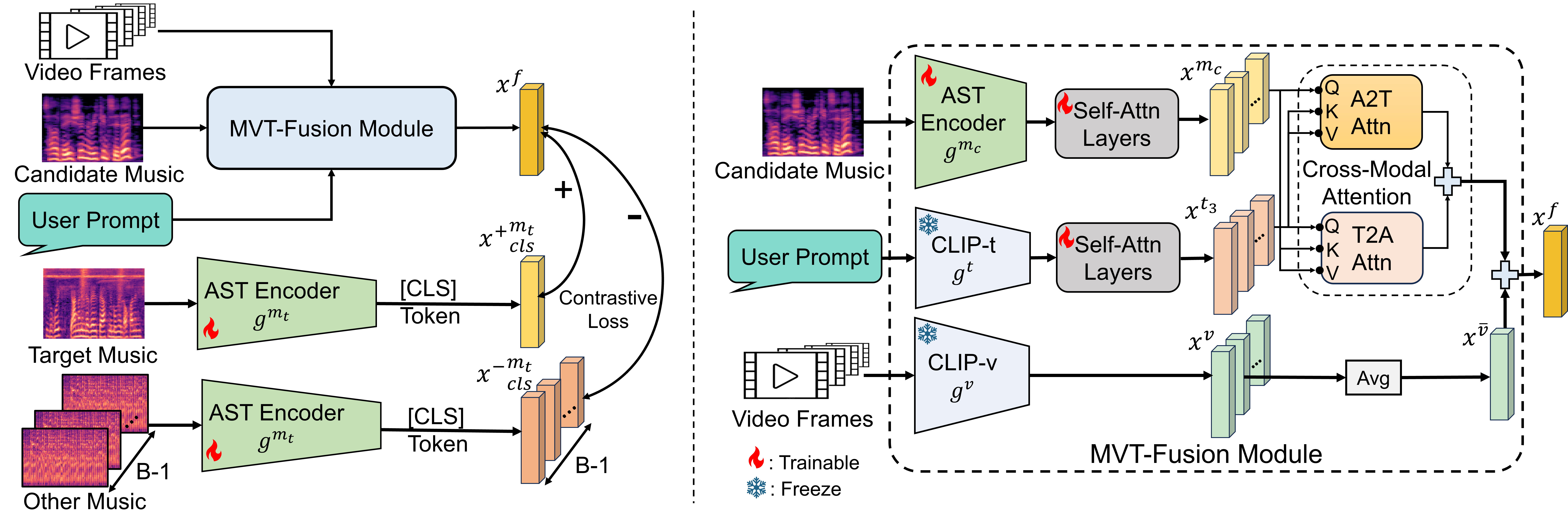}

   \caption{The music recommendation module combines video, the candidate music (1st round result), and user prompt (2nd round input) to retrieve a music in a common embedding space, trained using multi-modal contrastive loss (left). The candidate music corrected by user prompt along with the original video should result in a representation ``closer'' to the target music than other music. MVT-Fusion Module (right) is designed to combine the 3 modalities into an embedding space: i). The candidate music is encoded using the Audio Spectrogram Transformer (AST) \cite{gong2021ast}. ii). User prompt is fed into CLIP \cite{radford2021learning} text encoder (freeze) to get unpooled features. iii). Average pooling is performed on CLIP (freeze) vector of each video frame to obtain the video representation. iv). To foster fusion between candidate music and user prompt, self-attention layers and cross-modal attention (A2T and T2A, `T' - text, `A' - audio) are added to obtain the fused music-text vector.}
   \label{fig:ranking_module}
\end{figure*}
As illustrated in \Cref{fig:ranking_module}, we focus on enhancing the model's ability to transit the recommendation from the existing candidate music $m_c$ to target music $m_t$. Intuitively, when user prompt is given as an additional context to the candidate music and original video, the information combined from these modality contexts should arrive at a representation staying as close to the target music as possible. This motivates our formulation of the recommendation module as a metric learning problem: To learn a common embedding space between the tri-modal combination (candidate music + video + user prompt) and music. Towards this end, we propose a novel MVT-Fusion Module, as shown in \Cref{fig:ranking_module} (right), that explores music-text-video fusion in a fine-grained manner to obtain the tri-modal embedding. The embedding then serves as a goal embedding with which the target music is aligned. We utilize contrastive learning for the embedding alignment. We detail MVT-Fusion Module and contrastive formulation respectively.

\textbf{MVT-Fusion Module} takes the candidate music $m_c$ (represented as mel-spectrogram), user prompt $t_3$, and video frames $v$ as inputs. Each input is encoded via pretrained encoder in their corresponding modality. The candidate music is encoded using a AST (Audio-Spectorgram-Transformer)~\cite{gong2021ast} $g^{m_c}$. The user prompt is fed into CLIP~\cite{radford2021learning} text encoder $g^{t}$ to obtain contextualized language features (unpooled). Each video frame is transformed into a vector using CLIP image encoder $g^v$, and is then averaged along the temporal axis to obtain a single vector $\mathbf{x}^{\bar{v}}$, representing the semantics of the video.

To better capture the correlation between the candidate music and the user text prompt, we develop a fusion method for merging the features from music and text modalities. We adopt the ``late fusion'' strategy, applying several self-attention layers to both output features from the CLIP text encoder and the AST encoder before fusion. The output features from self-attention layers of two branches are $\mathbf{x}^{t_3} \in \mathcal{R}^{n_t \times d}$ and $\mathbf{x}^{m_c} \in \mathcal{R}^{n_m \times d}$, and expanded as:
\begin{equation}
\begin{aligned}
\mathbf{x}^{t_3} &= \left[x_{\text{cls}}^{t_3},  x_1^{t_3}, \ldots, x_{(n_t - 1)}^{t_3}\right], \\
\mathbf{x}^{m_c} &= \left[x_{\text{cls}}^{m_c},  x_1^{m_c}, \ldots, x_{(n_m - 1)}^{m_c}\right],
\end{aligned}
\end{equation}
where variable with subscript $cls$ serves as their summary of the respective sequence, along with the other elements capturing detailed features.

Then, we fuse the transformed features by implementing a cross-modal attention layer, which is defined as:
$\operatorname{Att}(Q, K, V)=\operatorname{softmax}\left(\frac{Q K^{\top}}{\sqrt{d_k}}\right) V$,
where $d_k$ is the dimensionality of key vectors. $Q$ and $K$, $V$ are from two different modalities. 

We add the video vector $\mathbf{x}^{\bar{v}}$ to obtain the tri-modal fusion vector representation $\mathbf{x}^f$:
\begin{equation}
\mathbf{x}^f = \mathbf{x}^{\Bar{v}} + \operatorname{Att}(x_{\text{cls}}^{t_3}, \mathbf{x}^{m_c}, \mathbf{x}^{m_c}) + \operatorname{Att}(x_{\text{cls}}^{m_c}, \mathbf{x}^{t_3}, \mathbf{x}^{t_3})
\label{eq:final_fusion}
\end{equation}

\textbf{Contrastive Formulation} Once $\mathbf{x}_f$ is obtained as the tri-modal fusion feature, it is necessary to align the representation of the target music with it. To achieve this, we first transform music from the pool into a vector space using a separate AST encoder $g^{m_t}$. Then we propose a contrastive learning approach to learn a common vector space between tri-modal fusion vector and music vector. Specifically, we use the hidden state of the last layer corresponding to the $cls$ token as the music vector. We denote the target music vector as $\mathbf{x}^{m^{cls}_t}_{+}$, and any other music vector as $\mathbf{x}^{m^{cls}_t}_{-}$. Then, we formulate a contrastive loss aiming at keeping the vector distance between $\mathbf{x}_f$ and $\mathbf{x}^{m^{cls}_t}_{+}$ closer while pushing away $\mathbf{x}^{m^{cls}_t}_{-}$ from any other music, as illustrated in \Cref{fig:ranking_module} (left). Specifically, we use the Contrastive Multiview Coding Loss \cite{tian2020contrastive}, a cross-modal variant of InfoNCE \cite{oord2018representation}. For each batch $B$, we have:
\begin{equation}
  \mathcal{L_R} = -\sum_{i=1}^B\left[\log \frac{h\left(\mathbf{x}^f_{(i)}, \mathbf{x}^{m^{cls}_t}_{(i)}\right)}{\sum_{j \neq i} h\left(\mathbf{x}^f_{(i)}, \mathbf{x}^{m^{cls}_t}_{(j)}\right)+h\left(\mathbf{x}^f_{(i)},  \mathbf{x}^{m^{cls}_t}_{(i)}\right)}\right] ,
  \label{eq:multiview_loss}
\end{equation}
where $\mathbf{x}^f_{(i)}$ and $\mathbf{x}^{m^{cls}_t}_{(i)}$ are $i$-th fusion vectors and target music vectors in the batch respectively. $h(\mathbf{x},\mathbf{y})=\exp\left(\frac{\mathbf{x}^\top\mathbf{y}}{\tau}\right)$ is a discriminating function, with $\tau$ being a trainable temperature hyperparameter.

\subsection{Sentence Generator}
To equip MuseChat with reasoning capability, we propose a sentence generator to provide justification for the recommended music. Towards this end, we design a multi-modal LLM, as illustrated in \Cref{fig:sentence_generator}, using Vicuna-7B \cite{zheng2023judging} (derived from fine-tuning Llama2-7B \cite{touvron2023llama} model) as the backbone. Each training instance comprises a music representation, denoted as $\mathbf{x}^{\bar{m}_t}$, which is the average of the music embedding $\mathbf{x}^{m_t} = [x_{\text{cls}}^{m_t},  x_1^{m_t}, \ldots, x_{(n_m - 1)}^{m_t}]$, derived from the previously fine-tuned AST Encoder $g^{m_t}$. The instance also includes the corresponding recommendation reasoning statement, $t_4$, from the simulated conversations. Notably, in our settings, each piece of music, whether as a candidate or target, has a corresponding reasoning statement in a conversation. Therefore, we do not specifically use the reasoning statement of candidate music in the first turn, $t_2$, because when this candidate music serves as the target in the other conversation, its reasoning statement is already utilized. We apply a linear layer $f_l$ to project the averaged music features $\mathbf{x}^{\bar{m}_t}$ onto the text embedding space. To increase training efficiency, we leverage LoRA \cite{hu2021lora} to fine-tune the attention and output layers of Vicuna. We use next token prediction task aiming at minimize the negative log-likelihood of response tokens conditioned on the recommended music:

\begin{figure}[!h]
  \centering
   \includegraphics[width=0.9\linewidth]{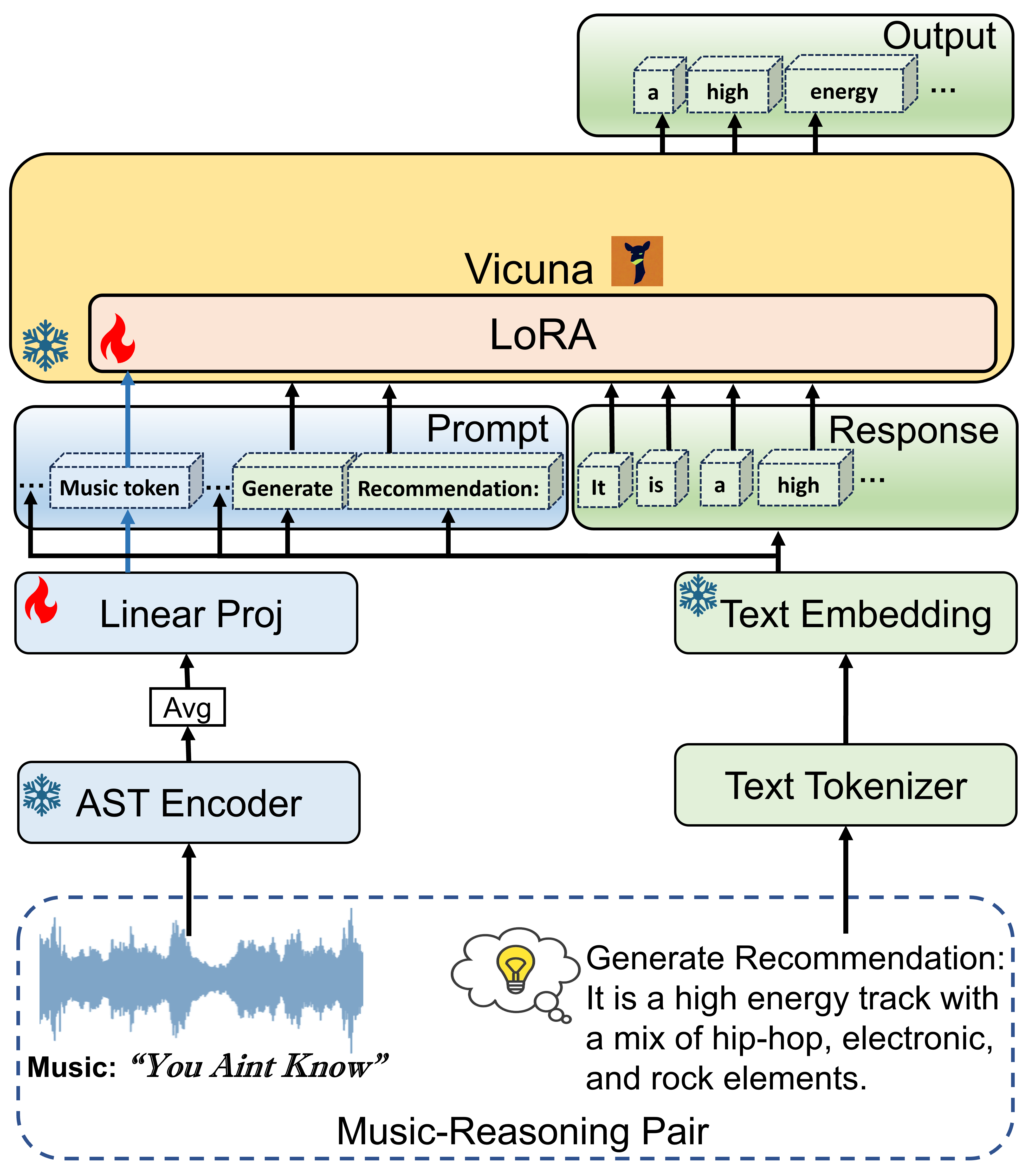}
   \caption{Illustration of sentence generator. During training, we only train the linear projection layer and the additional LoRA weights while keeping the parameters of Vicuna-7B and AST encoder frozen. And the prompt input is ``\textit{\#\#\# Recommender: Music feature: $<$Music$>$ [music token] $<$/Music$>$; Generate Recommendation:}".
   During inference, we give the recommended music title as additional information, and the prompt input is ``\textit{\#\#\# Recommender: Music title: [title]; Music feature: $<$Music$>$ [music token] $<$/Music$>$; Generate Recommendation:}". ``\textit{[music token]}" corresponds to the embedding vector projected from AST encoder output.
   }
   \label{fig:sentence_generator}
\end{figure}
\begin{equation}
\mathcal{L_G}\left(\mathbf{y} ; \mathbf{\theta}\right)=-\sum_{i=1}^n log\left[p_\theta\left(y_{i} \mid \left[f_l\left(\mathbf{x}^{\bar{m}_t}\right),y_{<i}\right]; \theta\right)\right],
  \label{eq:generation_loss}
\end{equation}
where $y_{i}$ is the $i$-th token in the response $\mathbf{y}$, and $\theta$ is the trainable parameters.

Although we train the sentence generator using only music embeddings, during inference, we also input the title of suggested music from the music recommendation module. This is necessary because the model cannot generate accurate names of unseen music tracks.

%% file: sec/results.tex
\section{Experimental Results}
\label{sec:experiments}
\subsection{Implementation Details}
We allocate 88,000 of music video clips to our training set. For each 120-second music video clip, we divide it into twelve 10-second segments and capture 5 frames per second from each segment. 
\begin{table}[!h]
  \small
  \centering
  \resizebox{1\columnwidth}{!}{%
  \begin{tabular}{@{}lccccc@{}}
    \toprule
    Model & MR $\downarrow$ & R@1 $\uparrow$ & R@5 $\uparrow$ & R@10 $\uparrow$ & SR $\uparrow$ \\
    \midrule
    Sum-Fusion & 17 & 10.58 & 28.47 & 40.19 & 21.70\\
    MVP & 7 & 20.71 & 48.89 & 63.14 & 20.71\\
    Self-Attn Fusion & 5 & 21.40 & 50.83 & 65.29 & 28.37\\
    Cross-Attn Fusion & 3 & 25.71 & 56.97 & 71.07 & 31.48\\
    MuseChat (ours) & \textbf{2} & \textbf{32.79} & \textbf{63.92} & \textbf{76.53} & \textbf{40.49}\\
    \midrule
    Chance & 250 & 0.20 & 1.00 & 2.00 & 0.40\\
    \bottomrule
  \end{tabular}%
  }
  \caption{Music retrieval results for different baseline models v.s MuseChat.}
  \label{tab:results_comparison}
\end{table}
In our training process for the music recommendation module, each training sample includes a 10-second video clip, a corresponding 10-second target music clip, a 10-second candidate music clip, and a user prompt. To extract video and text features, we use OpenAI's CLIP model. For audio features, we employ the AST model.\footnote{The pretrained weights used are clip-vit-base-patch32 for the CLIP Image encoder and MIT/ast-finetuned-audioset-10-10-0.4593 for the AST.} We project the extracted features from all modalities into 256-dimensional vectors using linear layers. We then apply 4 self-attention layers and 1 multi-head cross-attention layer (with 16 heads each) for fusing the candidate music and text features. For contrastive loss, we initialize temperature as 0.1, and set batch size to be 34 per GPU. For training, we use AdamW optimizer~\cite{loshchilov2017decoupled} with a learning rate of 4e-5 and a decay rate of 5e-4. In the sentence generator module, the maximum sequence length is set to 128, and the temperature is fixed to 0.1. We set the LoRA rank to 32 and train the module for 3 epochs using a learning rate of 5e-4 with linear decay. Training for both modules are conducted on 16 Nvidia V100 32G GPUs.
\subsection{Ranking Evaluation}
Following the setup for measuring music retrieval with free-form natural language tasks as described in \cite{mckee2023language}, we randomly create non-overlapping music pools using test data, each containing 500 tracks. Each pool has only one target music track for each video. For the track-level testing, we calculate embeddings for all 12 segments of each 120-second video and music track. We then take the average of these 12 embeddings to create a single representative embedding for each video and each music track. Using these averaged embeddings, we evaluate the performance of the music recommendation module. In the first turn, music is suggested based solely on video features, as we assume that the user does not provide any specific preferences at this point. In the second turn, we include the user's text prompt and candidate music along with the video features. This setup allows us to evaluate the system's ability to modify its initial recommendations based on the new information. For both turns, we rank music tracks by calculating the cosine similarity between the features of the music in the pool and the fusion features. We use various metrics such as Recall@K for K = 1, 5, 10, and Median Rank (MR) to evaluate the performance of the recommendation. We also measure the ``success rate" (abbreviated as SR), defined as the percentage of videos whose target music track appears at least once as the top of the recommended list within two turns. Since the MVP model cannot accommodate the user prompt as an input for the second turn, its SR is same as R@1.
We report the average performance for each of these metrics across all test music pools.

\textbf{Baselines Comparison} We introduce four baseline models with various structures and modalities to assess the effectiveness of MuseChat in ranking. (1) \textbf{MVP} model shares the same architecture as the MVP model for the data generation, but is trained on music videos from YouTube-8M than our proprietary data. (2) \textbf{Sum-Fusion} retains the architecture of MuseChat but employs a different fusion mechanism by summing up the vectors of the video, music, and text directly. We use the mean-pooled vector for both candidate music encoded by AST and text features extracted by CLIP text encoder. (3) \textbf{Self-Attn Fusion} extends the summation approach by including self-attention layers for text and music modalities before fusion, capturing intra-modal dynamics. (4) \textbf{Cross-Attn Fusion} removes self-attention layers and keep the cross-attention layer between music and text branch. 
We train the baseline models using the same music video pools and loss function as MuseChat. As shown in \Cref{tab:results_comparison}, MuseChat outperforms all baseline models. In particular, in R@1 metric, our model achieves a significant gain of \textbf{+7.0\%} against ``Cross-Attn Fusion'' and \textbf{+12.1\%} against the ``MVP'' model, showing the effectiveness of our fusion approach as well as the use of all three modalities as inputs. Notably, for models that could take video, candidate music and user prompt as inputs, we report the recall and MR for the second turn to show the effectiveness of our model when dealing with user preferences. The first turn results for these models are detailed in the supplementary materials. Additionally, we explore the contributions of various modalities to the retrieval performance in our model, with these specifics also included in the supplementary materials.

Although MuseChat shares similarities in experimental setup with the studies in \cite{mckee2023language, suris2022time}, we avoid direct comparisons due to no access to their models and data. Furthermore, the model in \cite{mckee2023language} is not designed to incorporate candidate music as input. Therefore, being the first to address this gap, our findings demonstrate robust performance and establish a novel benchmark for the field.

\subsection{Modality Ablation Studies}
To further show the necessities of combining all three modalities for retrieval, we conduct ablation studies by removing video, candidate music, and text branch in turn during training. When excluding candidate music or text features, we remove the related cross-attention layers, retaining only the self-attention layers, with the $cls$ token used for the summation of modality features. As \Cref{tab:modality_ablation} indicates, the model without visual branch, which is only valid for the second turn, exhibits the worst performance. This reveals the significance of visual information. Additionally, when comparing the MVP model with MuseChat without candidate music, we observe that user prompt helps to retrieve music more accurately.
\begin{table*}[!t]
  \centering
  \resizebox{0.95\linewidth}{!}{
  \begin{tabular}{@{}llccccr@{}}
    \toprule
    Model & Modality & MR $\downarrow$ & R@1 $\uparrow$ & R@5 $\uparrow$ & R@10 $\uparrow$& SR $\uparrow$\\
    \midrule
    MVP & Video $\rightarrow$ Music & 7 & 20.71 & 48.89 & 63.14 & 20.71 \\
    MuseChat w/o Video & (Music, Text) $\rightarrow$ Music & 19 & 8.12 & 24.53 & 37.11 & 8.12 \\
    MuseChat w/o Candidate Music & (Video, Text) $\rightarrow$ Music & 5 & 22.67 & 51.53 & 65.42 & 26.02 \\
    MuseChat (ours) & (Video, Music, Text) $\rightarrow$ Music & \textbf{2} & \textbf{32.79} & \textbf{63.92} & \textbf{76.53} & \textbf{40.49} \\
    \midrule
    Chance & - & 250 & 0.20 & 1.00 & 2.00 & 0.40 \\
    \bottomrule
  \end{tabular}
  }
  \caption{Ablation Studies: Performance comparison when removing modality branches from MuseChat and training corresponding models from scratch. }
  \label{tab:modality_ablation}
\end{table*}

\subsection{Reasoning Evaluation}
\begin{table*}[!h]
  \centering
  \resizebox{1.0\linewidth}{!}{
  \begin{tabular}{@{}llcccccr@{}}
    \toprule
    Model & Input Modality & BertScore \cite{zhang2020bertscore} (f1) $\uparrow$ & AB Divergence \cite{colombo2022infolm} $\downarrow$ & $\mathcal{L}_2$ Distance $\downarrow$ & Fisher-Rao Distance \cite{colombo2022infolm} $\downarrow$\\
    \midrule
    Vicuna-7B & Music Title & 0.9453 & 3.93 & 0.382 & 2.11 \\
    Vicuna w/ Music & Music Embeddings & 0.9526 & 2.68 & 0.279 & 2.02 \\
    MuseChat (ours) & Music Title + Embeddings & \textbf{0.9676} & \textbf{1.51} & \textbf{0.208} & \textbf{1.47} \\
    \bottomrule
  \end{tabular}
  }
  \caption{Comparison of semantic similarity between output and simulated conversations using various metrics. BERTScore \cite{zhang2020bertscore} assesses token-level similarity, while AB Divergence, $\mathcal{L}_2$ Distance, and Fisher-Rao Distance are derived based on InfoLM \cite{colombo2022infolm}.}
  \label{tab:reasoning_results_comparison}
\end{table*}
We introduce two baseline models to highlight the significance of training our sentence generator module using both music embeddings and music titles as inputs. The first baseline employs the frozen Vicuna-7B \cite{zheng2023judging} model, which is fine-tuned on the Llama2-7B \cite{touvron2023llama} weights. Since this model cannot process music embeddings, we only present it with the music title. The second baseline (Vicuna w/ Music) uses the same architecture as our sentence generator module but takes only music embeddings as input. We employ various common NLG metrics \cite{zhang2020bertscore, colombo2022infolm} to evaluate the performance of the models on simulated conversations. As shown in \Cref{tab:reasoning_results_comparison}, the Vicuna-7B model performs the worst. It struggles to capture the musicality of the given track, as it is a text-only model. As for Vicuna w/ Music, while capturing the musicality of the recommended track due to its training on both music and text, it fails to identify the correct music name and artist name based on audio information only. In contrast, by taking both audio information and music title as inputs, our sentence generator module achieves the best performance on reasoning.
\begin{figure}[!t]
  \centering
   \includegraphics[width=1\linewidth]{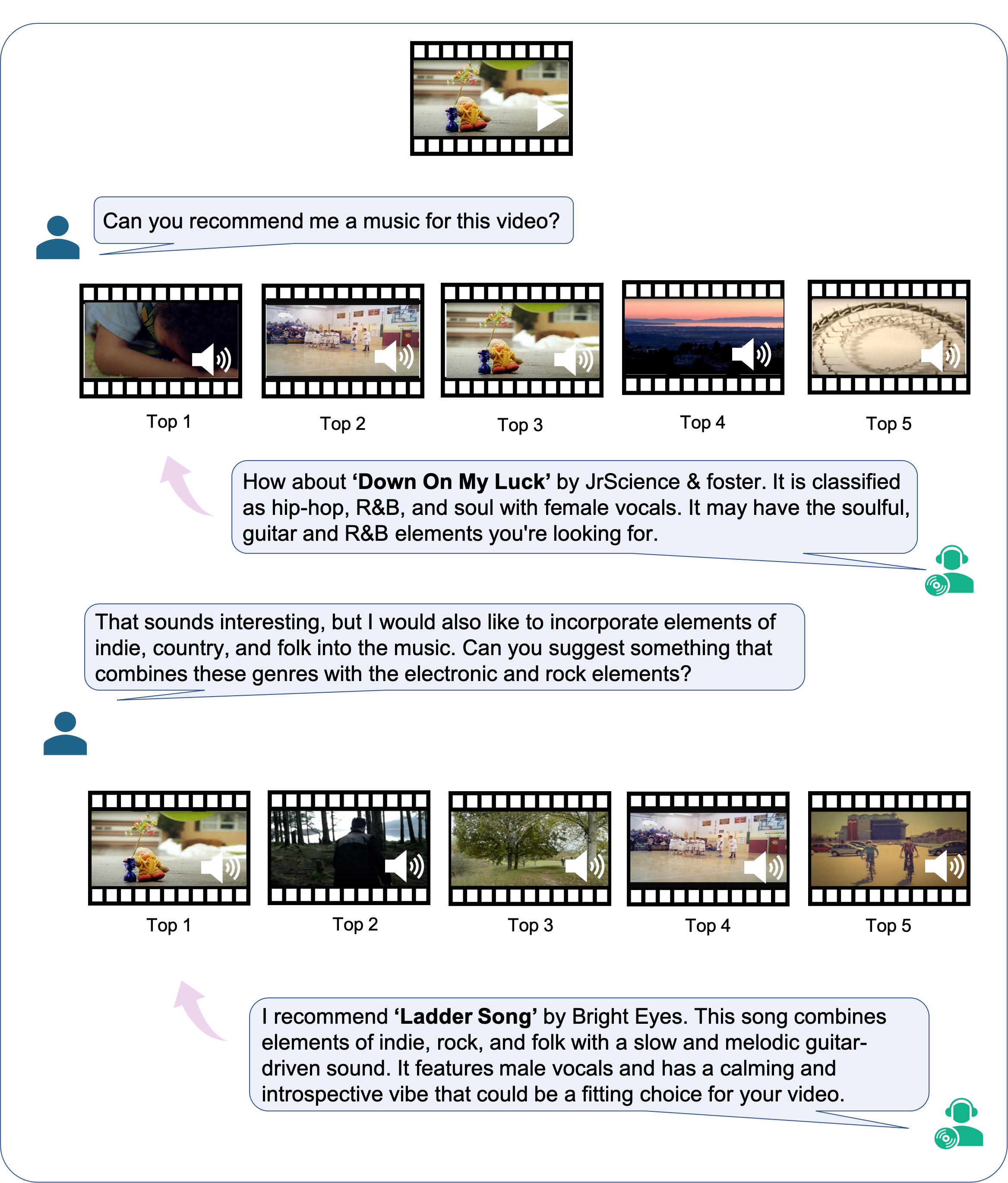}
   \caption{An example when MuseChat retrieves target music in the two turns.}
   \label{fig:qualitative_result}
   \vspace{-3mm}
\end{figure}

We conduct a qualitative evaluation of our MuseChat system to demonstrate its effectiveness in conversational music recommendations. This evaluation illustrates the system's ability to comprehend user queries, recommend relevant music, and produce responses that incorporate visual content, user preferences, and audio attributes. \Cref{fig:qualitative_result} shows how MuseChat interacts with users, dynamically adjusting its recommendations based on a range of contextual cues. We include more qualitative examples in the supplementary materials. 

Additionally, we assess our model and two baseline models through a human evaluation, using a 5-point MOS scale to measure correctness (identification of music and artist names), musicality (explanation of music characteristics), and clarity (understandability and coherence of the responses). \Cref{tab:human_eval} shows the human evaluation results. Since Vicuna-7B only takes the music title as input, it achieves a better score in correctness compared to Vicuna w/ Music, which struggles to identify the title and artist of an unknown piece of music based solely on music characteristics. In contrast, the Vicuna w/ Music model captures musicality better than Vicuna-7B, which relies solely on its knowledge base for musicality assessment. MuseChat, however, achieves the best overall performance by both correctly identifying music and artist name and understanding music in the context of video and user preference. Details about the human evaluation are included in the supplementary materials.
\begin{table}[!h]
  \footnotesize
  \centering
  \begin{tabular}{@{}lccccr@{}}
    \toprule
    Model & Correctness & Musicality & Clarity & Overall \\
    \midrule
    Vicuna-7B & 3.07 & 2.54 & 3.60 & 3.07 \\
    Vicuna w/ Music & 1.24 & 3.50 & 4.05 & 2.93 \\
    MuseChat (ours) & \textbf{4.63} & \textbf{4.22} & \textbf{4.54} & \textbf{4.46} \\
    \bottomrule
  \end{tabular}
  \caption{Human evaluation scores for music reasoning outputs.}
  \label{tab:human_eval}
  \vspace{-5mm}
\end{table}

%% file: sec/conclusions.tex
\section{Conclusions}
\label{sec:conclusion}
In this work, we establish a closer connection between human and music recommendation system through an interactive dialogue. We build the first conversational music recommendation dataset for videos based on public YouTube-8M dataset. In addition, we propose a system of two modules that interpret multi-modal inputs and deliver its recommendation with reasoning in natural language. Extensive experiments show that our proposed system exhibits strong performance on retrieval task with interpretability and interactivity. Future efforts could be made to design a more integrated system that combines recommendation and reasoning within a single multi-modal LLM.

%% file: sec/X_suppl.tex
\clearpage
\setcounter{page}{1}
\maketitlesupplementary

\section{Summary of Supplementary Materials:}
\noindent In this supplementary materials, we provide:
\begin{enumerate}
    \item Details of Our Conversational Music Recommendation dataset, including data distribution of music videos, prompt template and examples of simulated conversations, see Section ~\ref{sec:dataset_detail}.
    \item Details of the MVP model structure used for dataset construction (generate candidate music), see Section ~\ref{sec:mvp}.
    \item t-SNE visualization of music embeddings (from the recommendation module) associated with tags, see Section ~\ref{sec:t-SNE}.
    \item Additional Ablation Studies of the recommendation module, see Section ~\ref{sec:more_ablation}.
    \item Human Evaluation Details of the reasoning module, see Section ~\ref{sec:human_eval_details}.
    \item Qualitative Results that show the demo of our MuseChat, see Section ~\ref{sec:qual_results_conversation}.
\end{enumerate}

\section{Dataset Details}
\label{sec:dataset_detail}
In this section, we give details about our conversational music recommendation dataset including: (1) the distribution of music videos, and (2) the prompt template along with the examples of simulated conversations.

\subsection{Data Distribution}
\begin{figure}[!h]
  \centering
   \includegraphics[width=1\linewidth]{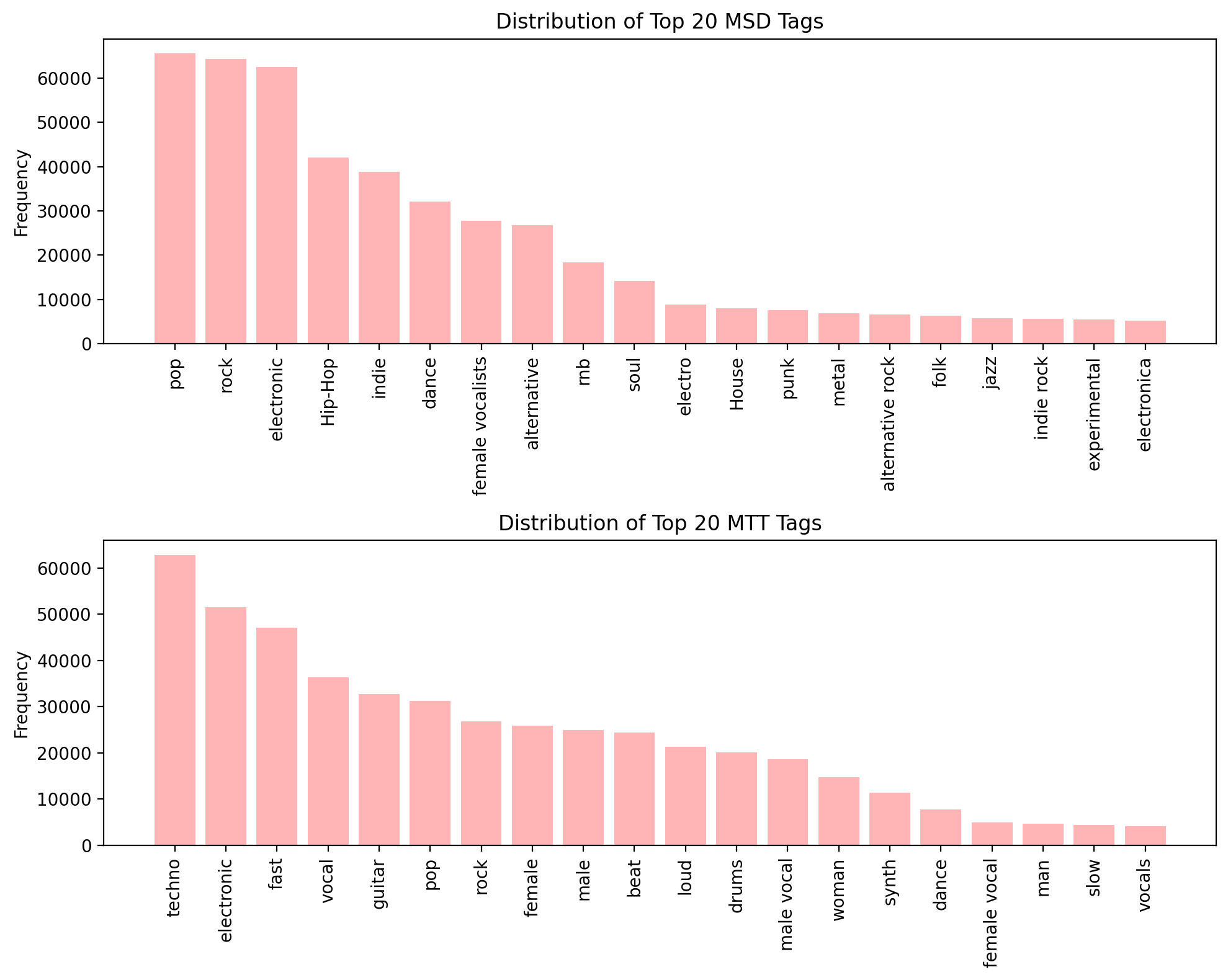}
   \caption{The distributions of the top 20 tags from the MSD~\cite{Bertin-Mahieux2011} and MTT~\cite{güçlü2016brains} systems used in our dataset.}
   \label{fig:tag_dist}
\end{figure}

To visualize the distribution of the music videos in the dataset, we plot top 20 music tags appeared in our dataset from MSD~\cite{Bertin-Mahieux2011} dataset and MTT~\cite{güçlü2016brains} dataset respectively. The distributions are shown in ~\Cref{fig:tag_dist}. 

\begin{figure*}[!h]
  \centering
   \includegraphics[width=0.9\linewidth]{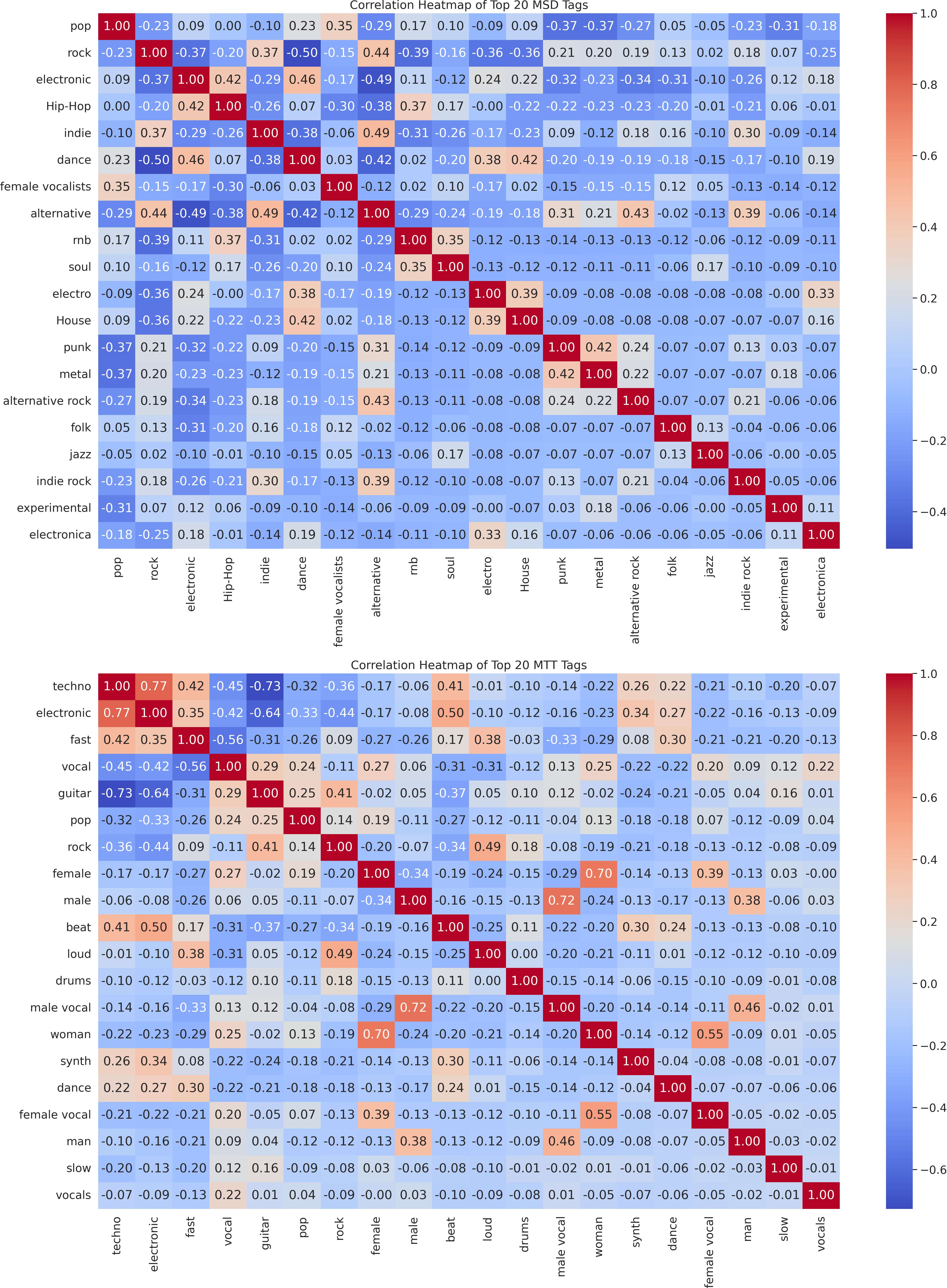}
   \caption{The correlation matrix of the top 20 tags from the MSD~\cite{Bertin-Mahieux2011} and MTT~\cite{güçlü2016brains} systems used in our dataset.}
   \label{fig:tag_corr}
\end{figure*}

To show the diversity of the applied tagging systems, we also study their correlation among top music tags, as shown in ~\Cref{fig:tag_corr}. From this figure, we observe that similar tags have higher correlation coefficient. For example, Hip-Hop music is highly correlated with electronic music, but is less correlated with folk music. And the widely spread ranges of the coefficients show that our tagging systems are capable of capturing a diverse range of music elements for different music tracks.

\subsection{Prompt Template and Simulated Conversations}
\label{sec:temp_sim}
\begin{figure*}[t]
    \centering
    \includegraphics[width=0.9\linewidth]{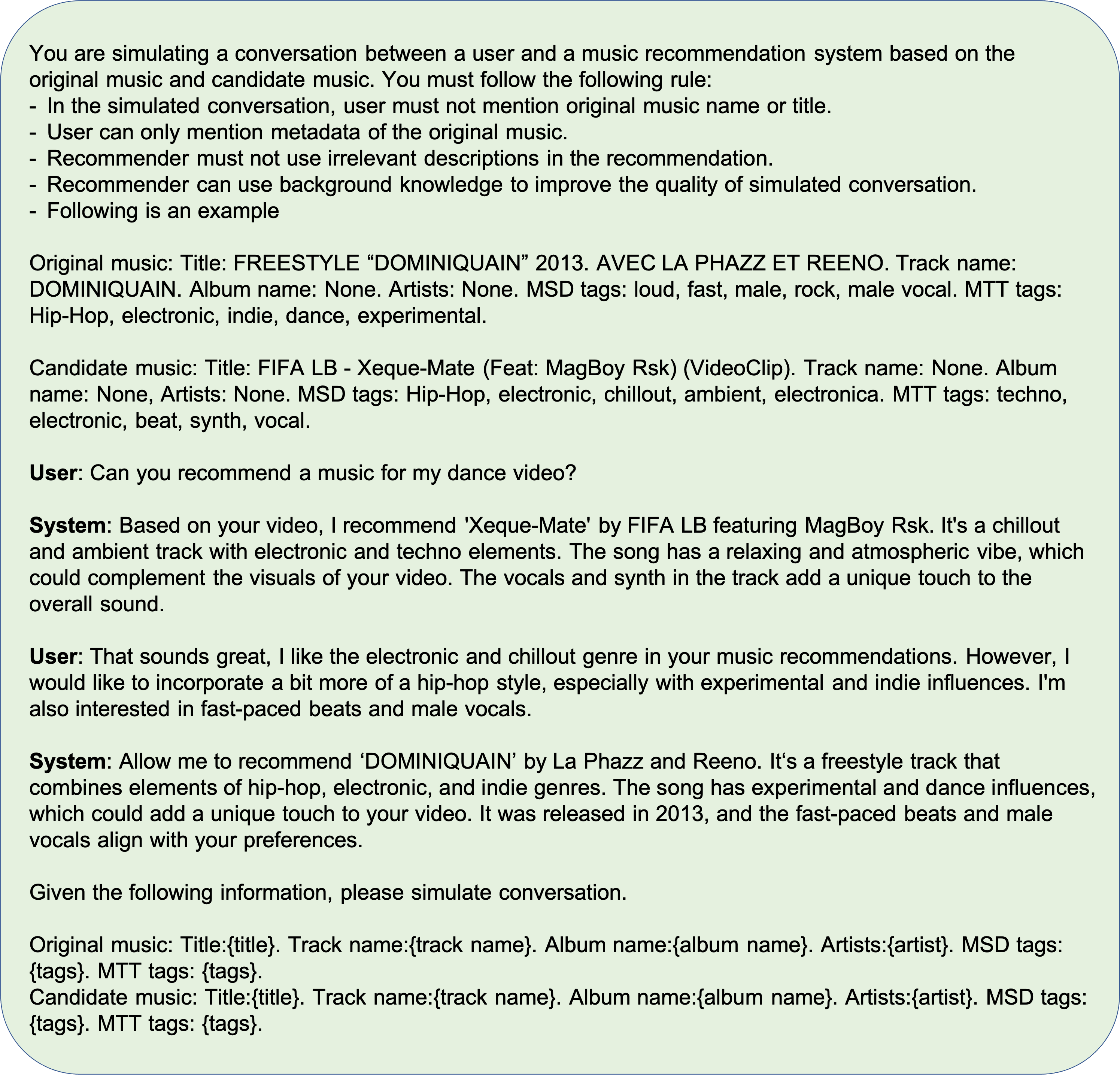}
    \caption{Prompt Template given to GPT-3.5 for dialogue simulation}
    \label{fig:prompt}
\end{figure*}

\begin{figure*}[t]
    \centering
    \includegraphics[width=0.85\linewidth]{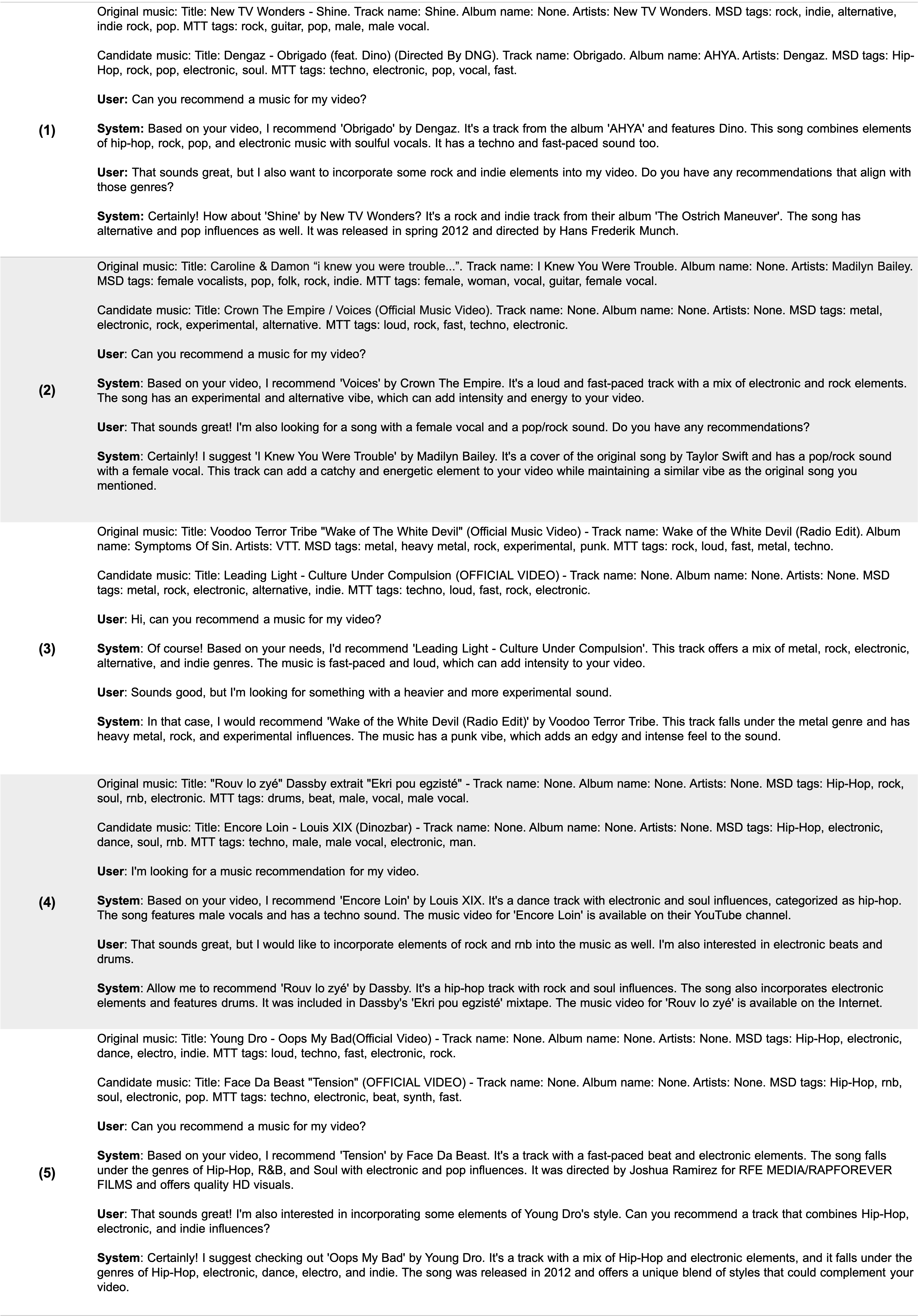}
    \caption{Examples of simulated conversations}
    \label{fig:example_conversation1}
\end{figure*}

\begin{figure*}[t]
    \centering
    \includegraphics[width=0.85\linewidth]{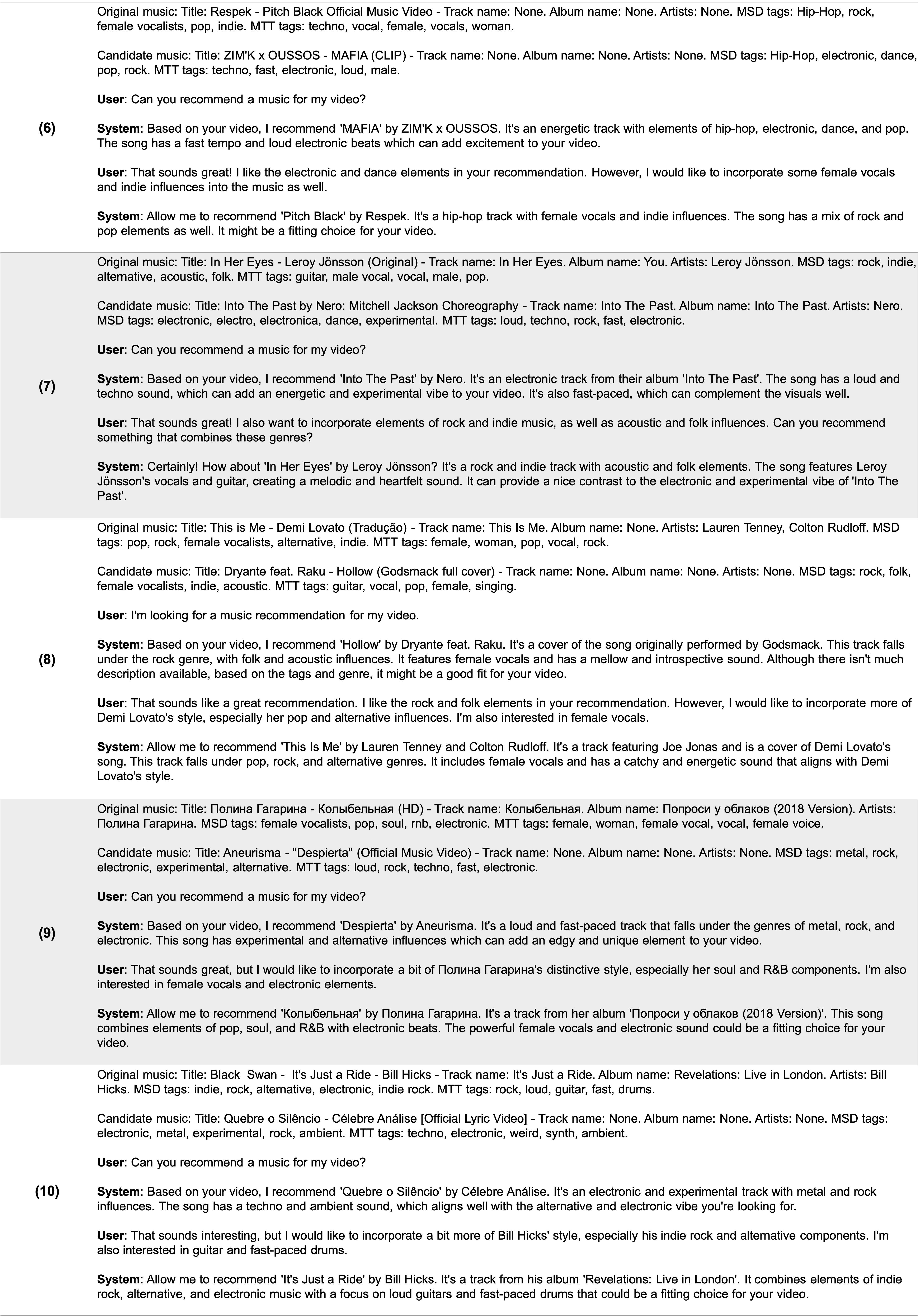}
    \caption{Examples of simulated conversations. Continued from \Cref{fig:example_conversation1}}
    \label{fig:example_conversation2}
\end{figure*}

\Cref{fig:prompt} illustrates the prompt template sent to GPT-3.5 for simulating conversations between a user and a recommendation system. Initially, we establish constrained rules to guide the generation of GPT-3.5. We then supply titles and top 5 music tags from each of two referenced datasets: the MagnaTagATune (MTT) dataset \cite{güçlü2016brains} and the Million Song Dataset (MSD) \cite{Bertin-Mahieux2011}. These tags apply to both original music and candidate music examples. If metadata like official track names, album names, or artist names are available, they are also included in the prompt. Finally, we provide human-written conversation templates featuring original and candidate music examples. During generation, we input different pairs of original and candidate music, guiding GPT-3.5 to create new conversations based on the provided human-written examples. \Cref{fig:example_conversation1} and \Cref{fig:example_conversation2} show 10 examples of simulated conversations based on the prompt template shown in \Cref{fig:prompt}.

\section{MVP structure}
\label{sec:mvp}
The Music Video Pretrained (MVP) model is structured as a two-tower system, sharing a similar architecture with the models described in \cite{huang2022mulan, suris2022time}. This model was trained on a dataset comprising 3 million pairs of short internal music videos in our proprietary dataset. These high-quality music video pairs, created by professionals or popular influencers, cover various topics such as dance, food/cooking, travel, vlogs, and sports. The MVP model extracts a 10-second segment from the input video and captures 5 frames per second from this segment. Similarly, the MVP extracts a 10-second segment from the input music and samples the audio signal at a rate of 12,000 Hz. For video processing, the model employs the CLIP Image encoder \cite{radford2021learning}. In terms of audio processing, the MVP utilizes an Audio Spectrogram Transformer (AST) \cite{gong2021ast} to extract music embeddings. Both video and music embeddings are then projected into a shared 256-dimensional latent space using respective trainable linear layers. During training, the MVP model employs the Contrastive Multiview Coding Loss, aiming to increase the distance between mismatched video-music pairs (negative pairs) and decrease it for well-matched pairs (positive pairs). Upon completion of its training, the MVP model computes the similarity between video and music segments to select the most fitting music for a given video.

\section{t-SNE visualizations}
\label{sec:t-SNE}
\begin{figure*}[!ht]
  \centering
  \includegraphics[width=0.9\linewidth]{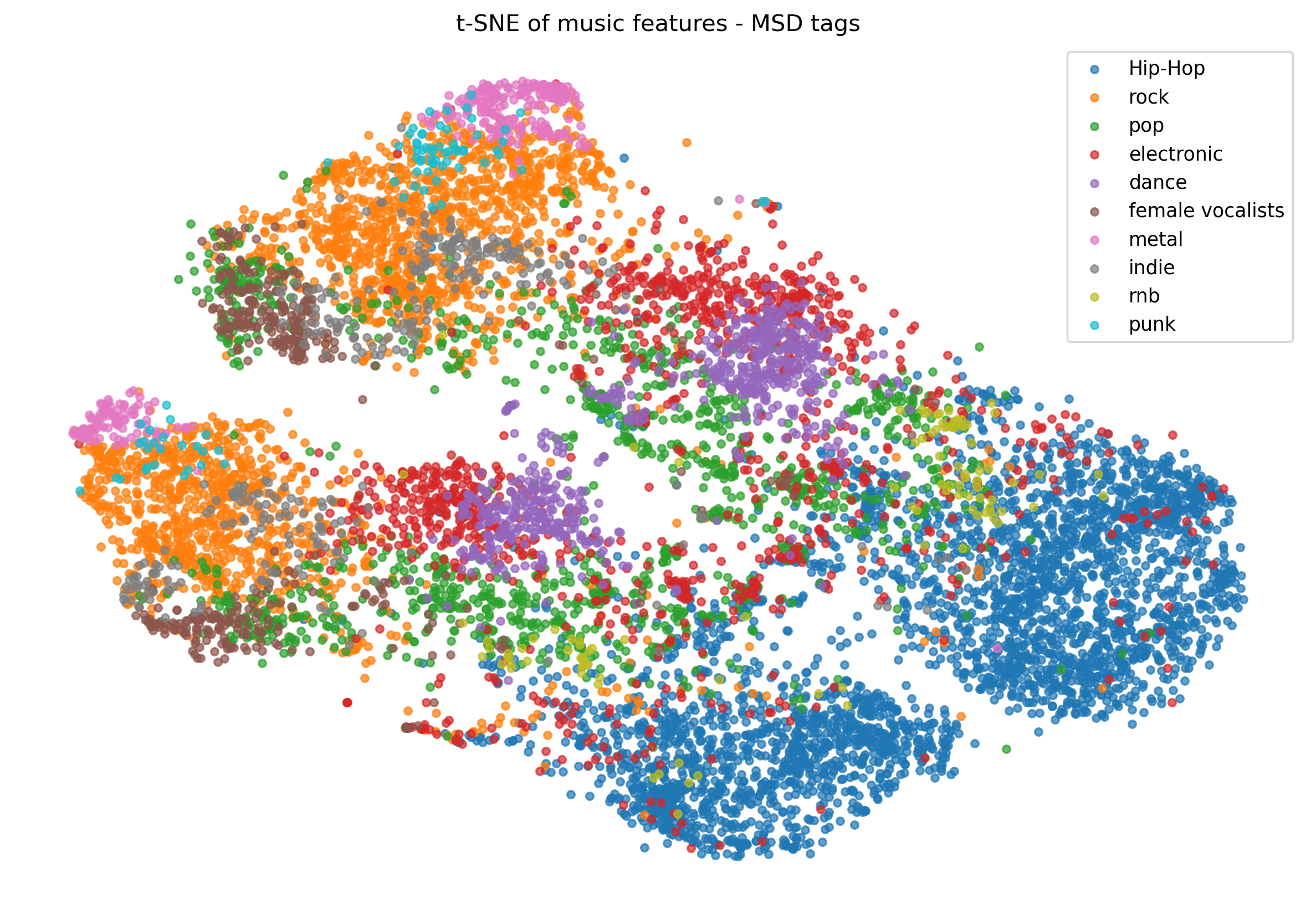}
  \includegraphics[width=0.9\linewidth]{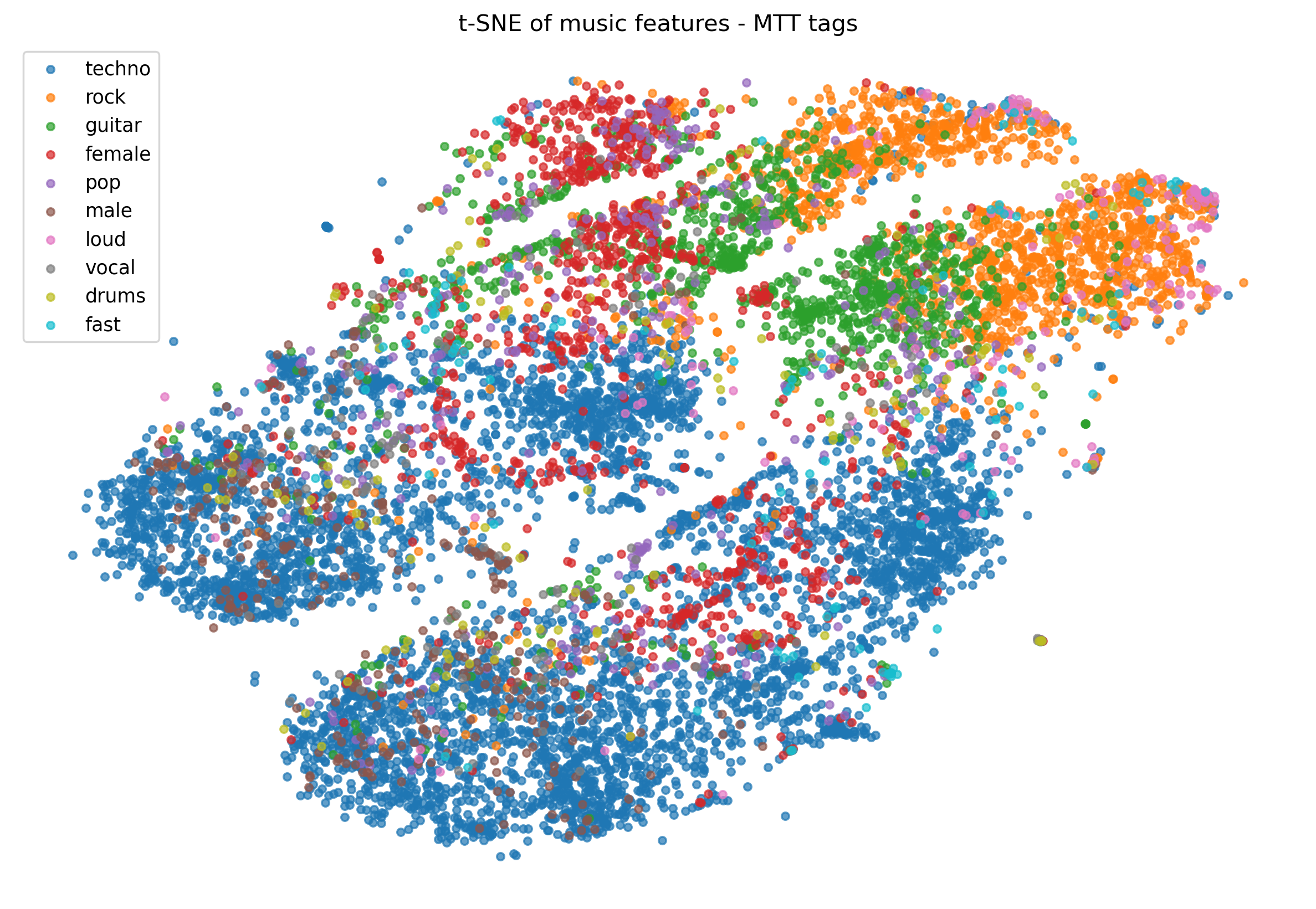}
  \caption{t-SNE visualizations of music features in MSD~\cite{Bertin-Mahieux2011} and MTT~\cite{güçlü2016brains} tags.}
  \label{fig:tsne}
\end{figure*}

To further study the quality of music embeddings learned from MVT-Fusion module, we use t-SNE to project the original music features, extracted using the AST branch, from our conversational music recommendation dataset into a 2-dimensional space.
We use the tag with highest probability from method \cite{pons2019musicnn} as the tag of the music track. For each music track in the test set, we extract the corresponding music feature using AST.  As shown in the plots of both MSD and MTT tags in \Cref{fig:tsne}, the music embedding is reasonably separated according to the tags.

\section{More Ablation Studies}
\label{sec:more_ablation}
We include additional ablation studies to manifest the effectiveness of the design choices in our proposed MVT-Fusion module.
\subsection{First Turn Retrieval Results}
In the first dialogue turn, we assume that user uploads video only for music recommendation. To demonstrate the capability of our proposed MVT-Fusion module in retrieving the music initially without user prompt as inputs, we compare different fusion baseline methods against our method. The results are shown in ~\Cref{tab:first_turn_result}. While all of these models are trained on three modalities -- video, candidate music and text, they achieve similar good results in music retrieval solely based on video as input. It is note-worthy that such tri-modalities models (MR 5) even outperform the MVP model (MR 7) with only video and music (Table 1 in main text).
\begin{table}[!h]
  \centering
  \begin{tabular}{@{}lcccc@{}}
    \toprule
    Module & MR $\downarrow$ & R@1 $\uparrow$ & R@5 $\uparrow$ & R@10 $\uparrow$\\ 
    \midrule
    Sum-Fusion & 5 & 21.70 & 49.49 & 63.27 \\ 
    Self-Attn Fusion & 5 & 21.44 & 50.30 & 64.09 \\ 
    Cross-Attn Fusion & 5 & 20.74 & 48.83 & 63.10 \\ 
    MuseChat (ours) & 5 & 20.74 & 48.83 & 63.10 \\ 
    \midrule
    Chance & 250 & 0.20 & 1.00 & 2.00 \\ 
    \bottomrule
  \end{tabular}
  \caption{First Turn Retrieval Performance using different fusion methods.}
  \label{tab:first_turn_result}
\end{table}
\subsection{Impact of Input Modality at the Test Stage}
We also examine how each modality contributes to the performance of MuseChat recommendation module. At the test time, we test two variants: (1) Use video as the only input, (2) Use candidate music along with user prompt text as inputs without addition of the video feature. As shown in ~\Cref{tab:input_modality_impact}, the model without the video feature perform worst, suggesting visual information as the most significant role to the retrieval performance. Moreover, when video information is supplemented with candidate music along with user prompt, the retrieval performance gets boosted further, matching both the context of the video and the user's preference.
\begin{table}[!h]
  \centering
  \resizebox{1\linewidth}{!}{
  \begin{tabular}{@{}lcccc@{}}
    \toprule
    Model & MR $\downarrow$ & R@1 $\uparrow$ & R@5 $\uparrow$ & R@10 $\uparrow$ \\
    \midrule
    MuseChat (music+text) & 22 & 7.37 & 23.16 & 34.04\\
    MuseChat (video) & 5 & 20.74 & 48.83 & 63.10\\
    MuseChat (ours) & \textbf{2} & \textbf{32.79} & \textbf{63.92} & \textbf{76.53}\\
    \midrule
    Chance & 250 & 0.20 & 1.00 & 2.00\\
    \bottomrule
  \end{tabular}
  }
  \caption{Impact of modality inputs on MuseChat at the test stage.}
  \label{tab:input_modality_impact}
\end{table}

\subsection{Modality Fusion Order Ablation}
In our proposed MVT-Fusion module, candidate music and user prompt text are fused before combined with the video feature. To demonstrate the efficacy of this strategy in terms of the fusion order, we explore two alternative fusion orders: (1) Integrating candidate music with video with cross-modal attention followed by addition to text, (2) Merging text with video with cross-modal attention followed by addition to candidate music. Both alternatives keep the same internal architecture as the MVT-Fusion module, with the order of fusion being the only difference. As shown in \Cref{tab:modality_fusion_order_ablation}, MuseChat's module outperforms the other two methods on both MR and recall metrics. We believe that the success of our strategy is attributed to the effective separation of visual information as an independent branch, reflecting its importance of contribution in the retrieval.

\begin{table}[!h]
  \centering
  \resizebox{1.0\linewidth}{!}{
  \begin{tabular}{@{}lccccc@{}}
    \toprule
    Model & MR $\downarrow$ & R@1 $\uparrow$ & R@5 $\uparrow$ & R@10 $\uparrow$ & SR $\uparrow$ \\
    \midrule
    Music-Video Fusion & 3 & 27.97 & 59.14 & 72.55 & 30.21\\
    Text-Video Fusion & 4 & 25.96 & 57.16 & 71.05 & 26.05\\
    MuseChat (ours) & \textbf{2} & \textbf{32.79} & \textbf{63.92} & \textbf{76.53}& \textbf{40.49}\\
    \midrule
    Chance & 250 & 0.20 & 1.00 & 2.00 & 0.40\\
    \bottomrule
  \end{tabular}
  }
  \caption{Ablation studies on fusion strategies in terms of fusion order}
  \label{tab:modality_fusion_order_ablation}
\end{table}

\section{Human Evaluation Details of the Reasoning Module}
\label{sec:human_eval_details}

\begin{figure}[t]
    \centering
    \includegraphics[width=1.0\linewidth]{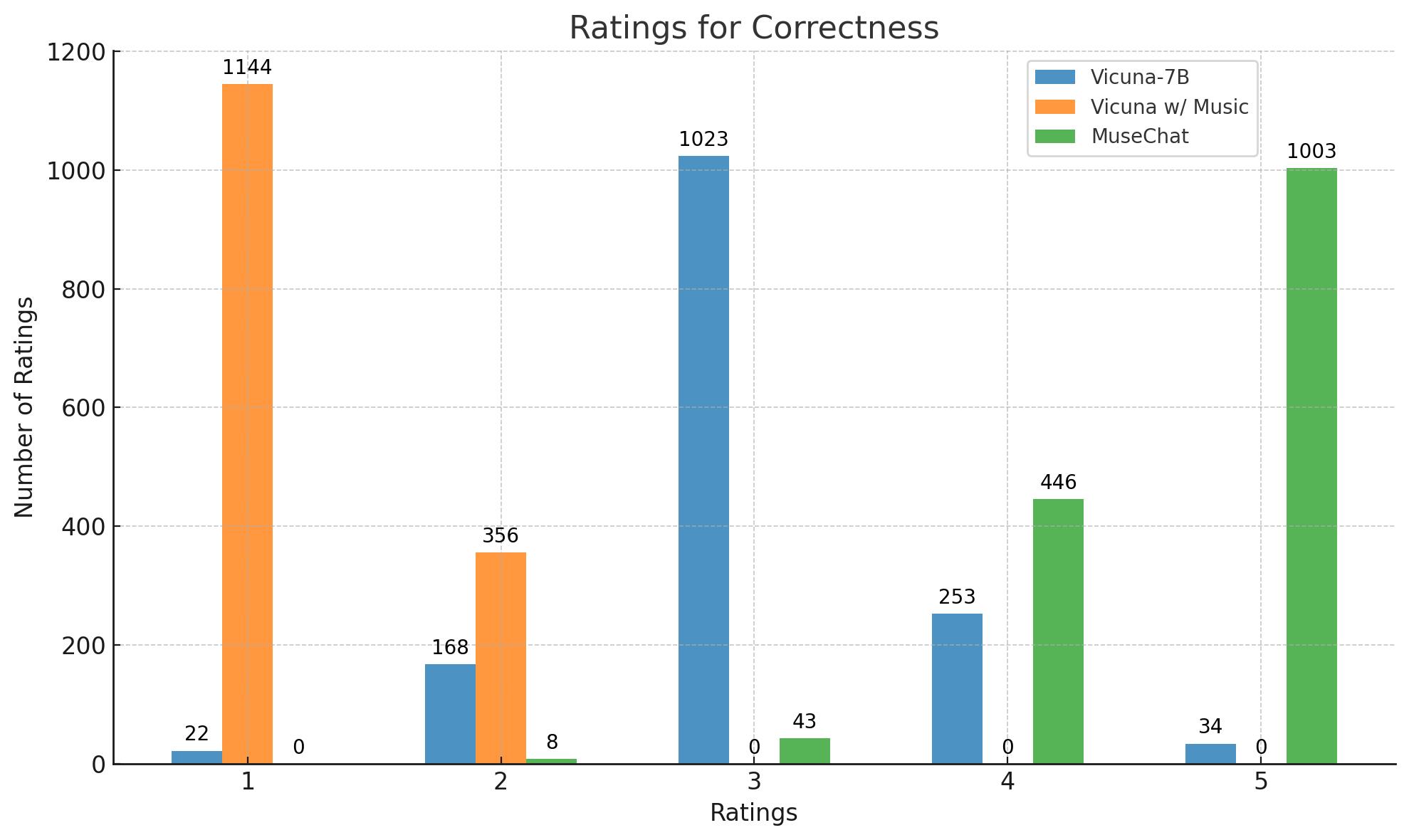}
    \caption{Human Evaluation Results for Correctness Aspect of Reasoning Module in MuseChat.}
    \label{fig:correctness}
\end{figure}

\begin{figure}[t]
    \centering
    \includegraphics[width=1.0\linewidth]{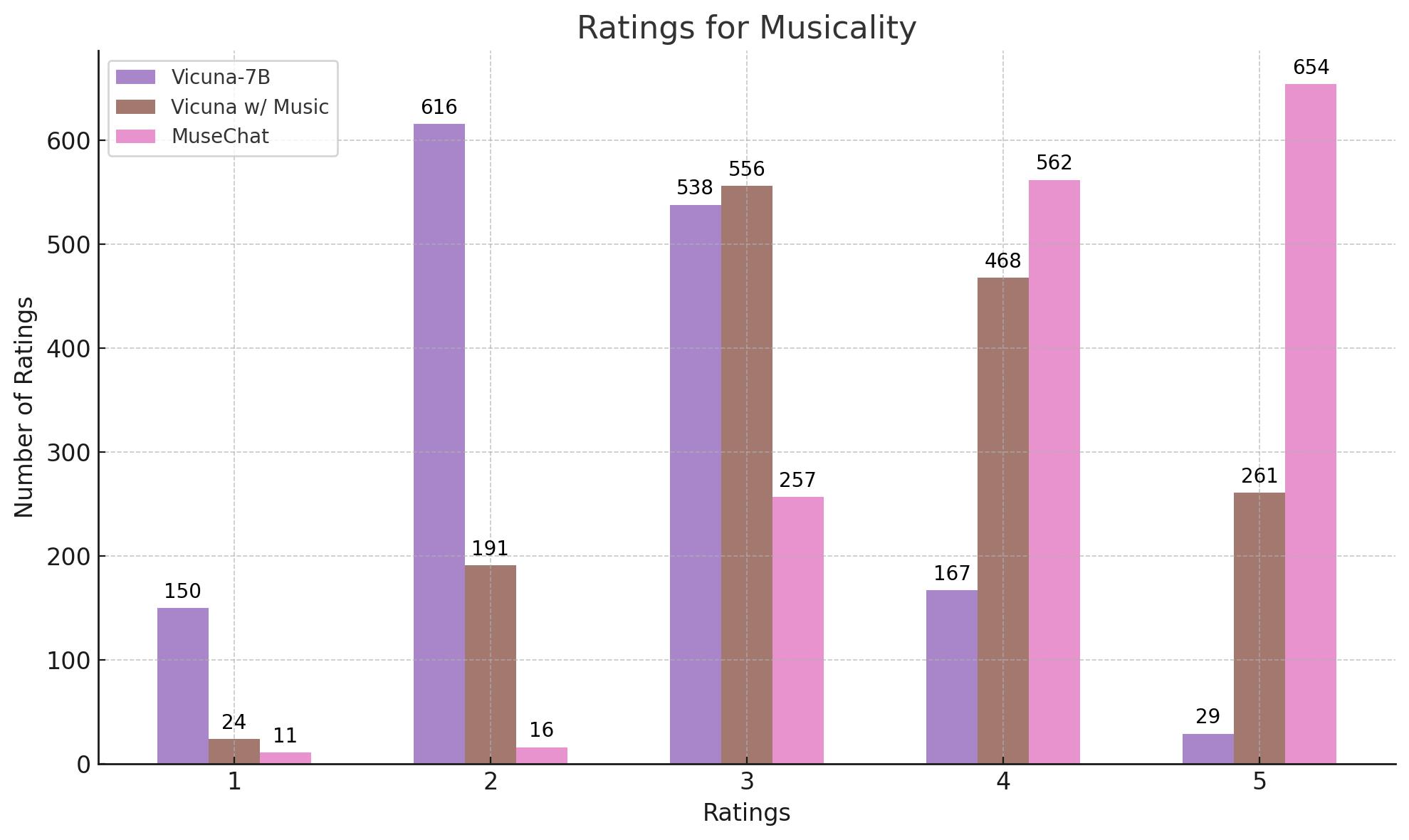}
    \caption{Human Evaluation Results for Musicality Aspect of Reasoning Module in MuseChat.}
    \label{fig:musicality}
\end{figure}

\begin{figure}[t]
    \centering
    \includegraphics[width=1.0\linewidth]{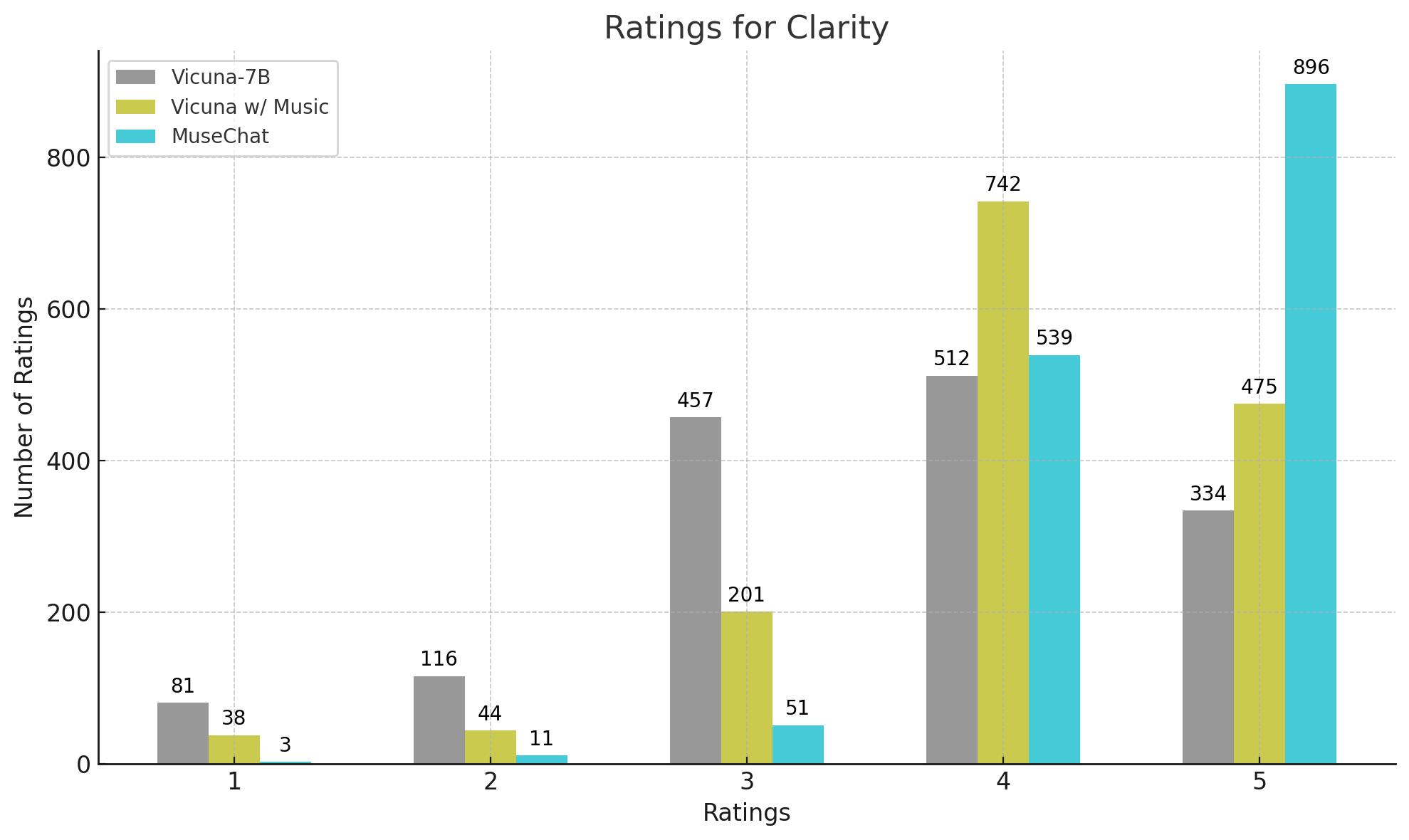}
    \caption{Human Evaluation Results for Clarity Aspect of Reasoning Module in MuseChat.}
    \label{fig:clarity}
\end{figure}

We use 5-point MOS (Mean-Opinion-Score) scale to measure correctness, musicality and clarity of the reasoning results. We include the following questions:
\begin{itemize}
    \item Correctness: How accurately does the system identify the music information?
    \begin{itemize}
        \item 1: Provides incorrect or irrelevant information about the music.
        \item 2: Identifies basic information but with significant inaccuracies.
        \item 3: Generally correct in identifying basic information, though some inaccuracies are present.
        \item 4: Accurately identifies most music information with minor errors.
        \item 5: Accurately and consistently identifies all relevant music information.
    \end{itemize}

    \item Musicality: How well does the system describe the music's characteristics in terms of depth and insight?
    \begin{itemize}
        \item 1: Incorrect explanation of the music's characteristics.
        \item 2: Basic explanation, lacking depth.
        \item 3: Adequate, covering essential characteristics with some depth.
        \item 4: Comprehensive explanation with considerable depth.
        \item 5: Exceptionally insightful and detailed explanation, capturing the essence of the music's characteristics.
    \end{itemize}

    \item Clarity: How well does the system convey its reasoning in terms of clarity, coherence, and completeness?
    \begin{itemize}
        \item 1: Outputs are generally incomplete or in the wrong format.
        \item 2: Outputs are mostly incomplete with some correct format elements.
        \item 3: Outputs convey main information but are not fully complete.
        \item 4: Outputs are mostly complete with minor errors or format issues.
        \item 5: Outputs are consistently complete, clear, and correctly formatted.
    \end{itemize}
\end{itemize}
Participants use the information from the webpage of corresponding YouTube video as the ground truth for correctness evaluation, listen to the retrieved music and compare with the reasoning text response to evaluate musicality, and assess the clarity of the response text as a whole.
To ensure robustness in our evaluation, we select 30 participants with different levels of music backgrounds from music lovers to music performers with years of experience. In the survey, participants are presented with a multiple-choice question to assess their musical skill level. They are asked to select from one of three options: (A) Music lover and show familiarity with the task; (B) Music practitioner with basic knowledge of music along with practical experience; and (C) Music expert who is experienced in at least one type of musical performance. The reported distribution of skill level among the participants is: (A) 12, (B) 11 and (C) 7. Such diversity of skill levels among the participants enhances the overall robustness of the final evaluation. To reduce the variance of the survey thus ensuring the quality, each participant is asked to review 50 reasoning text responses from each model: Vicuna-7B, Vicuna w/ Music, and MuseChat, focusing on assessing model performance in terms of correctness, musicality, and clarity. We show the distribution of ratings respectively in \Cref{fig:correctness,fig:musicality,fig:clarity}.

\section{Qualitative Results}
\label{sec:qual_results_conversation}
\begin{figure*}[t]
    \centering
    \includegraphics[width=0.8\linewidth]{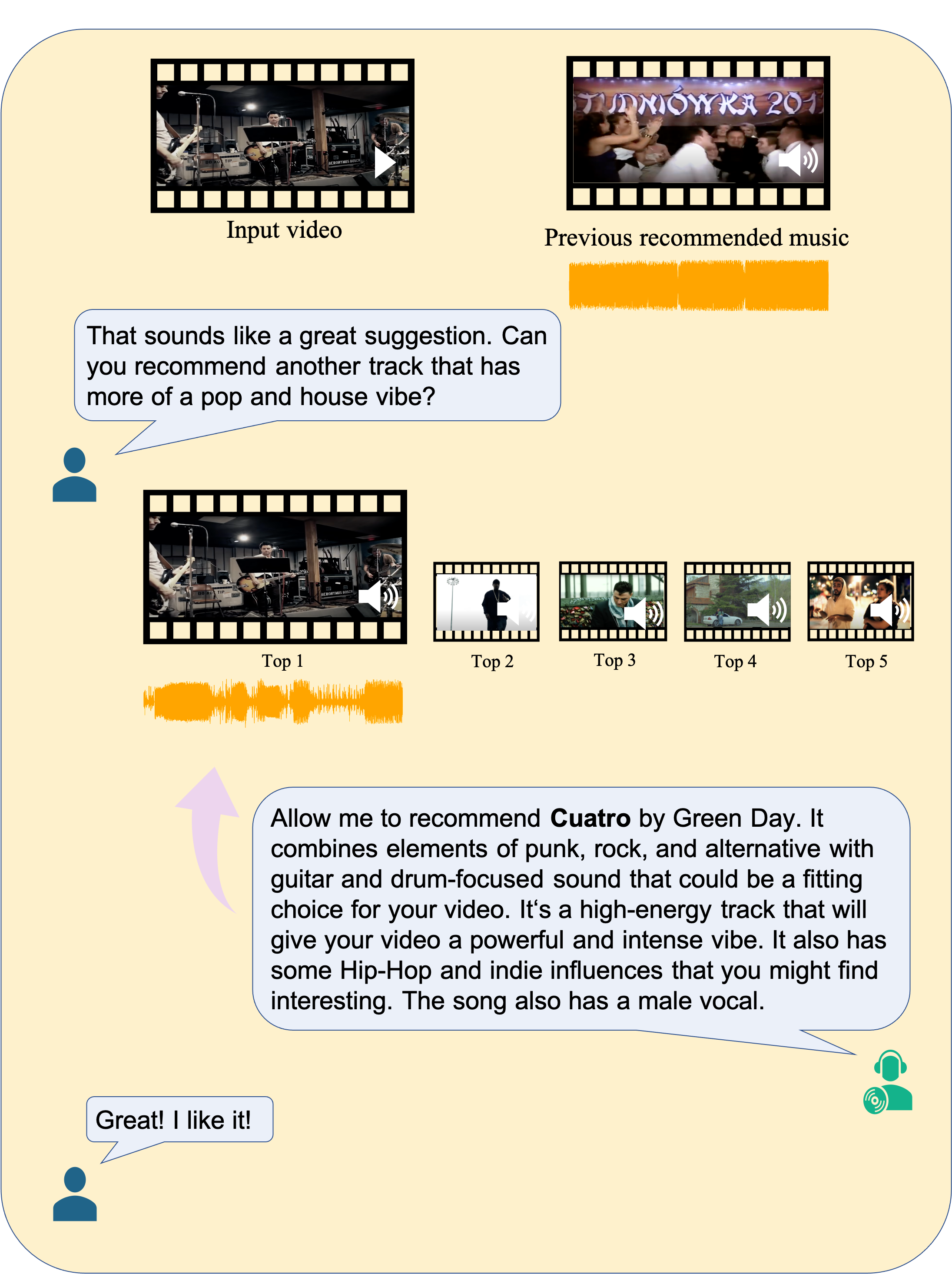}
    \caption{Example showing MuseChat's capability in improving recommendation results by learning user's preferences, contextual music and video content.}
    \label{fig:second_turn1}
\end{figure*}

\begin{figure*}[t]
    \centering
    \includegraphics[width=0.8\linewidth]{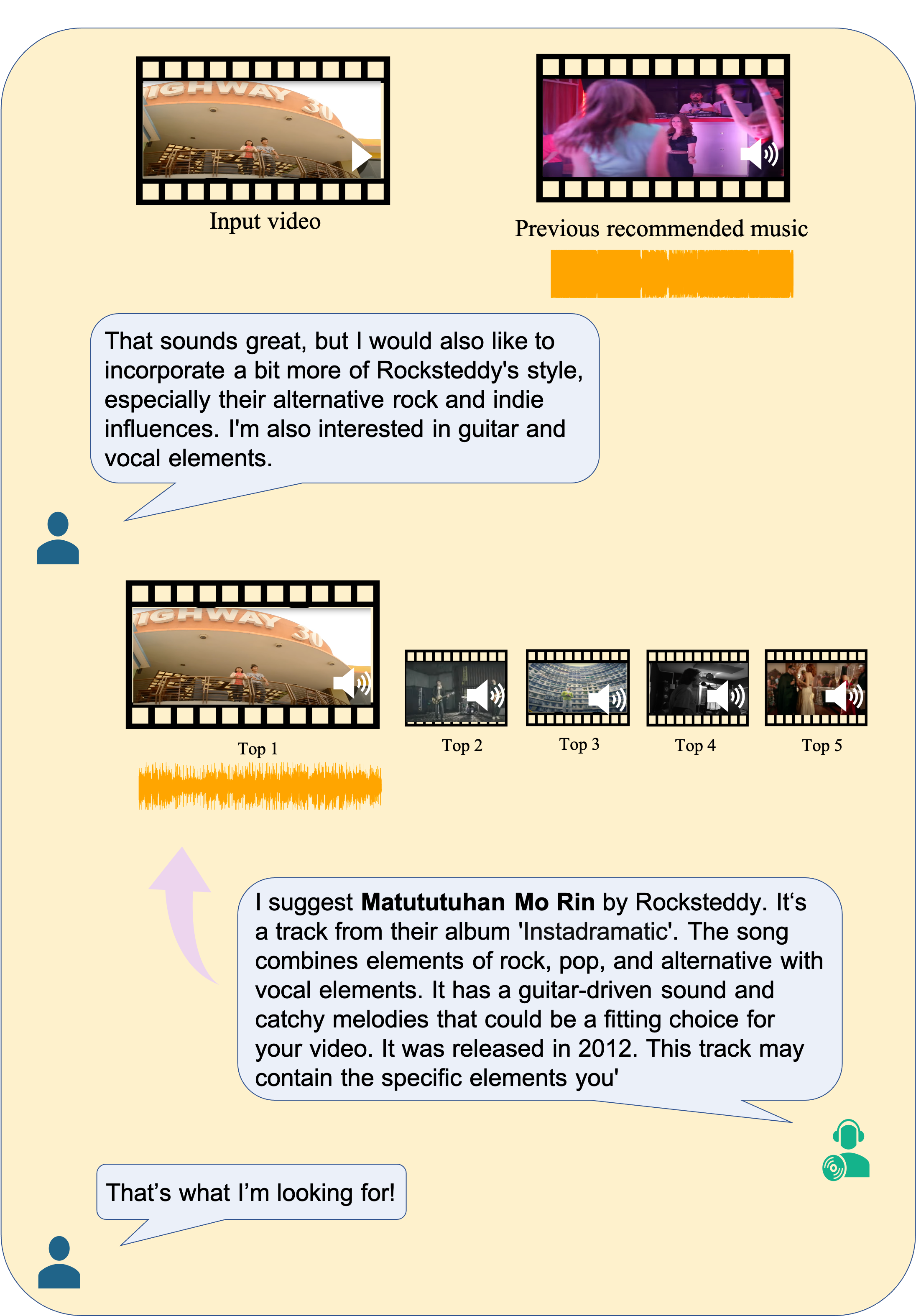}
    \caption{Example showing MuseChat's capability in improving recommendation results by learning user's preferences, contextual music and video content.}
    \label{fig:second_turn2}
\end{figure*}

\begin{figure*}[t]
    \centering
    \includegraphics[width=0.8\linewidth]{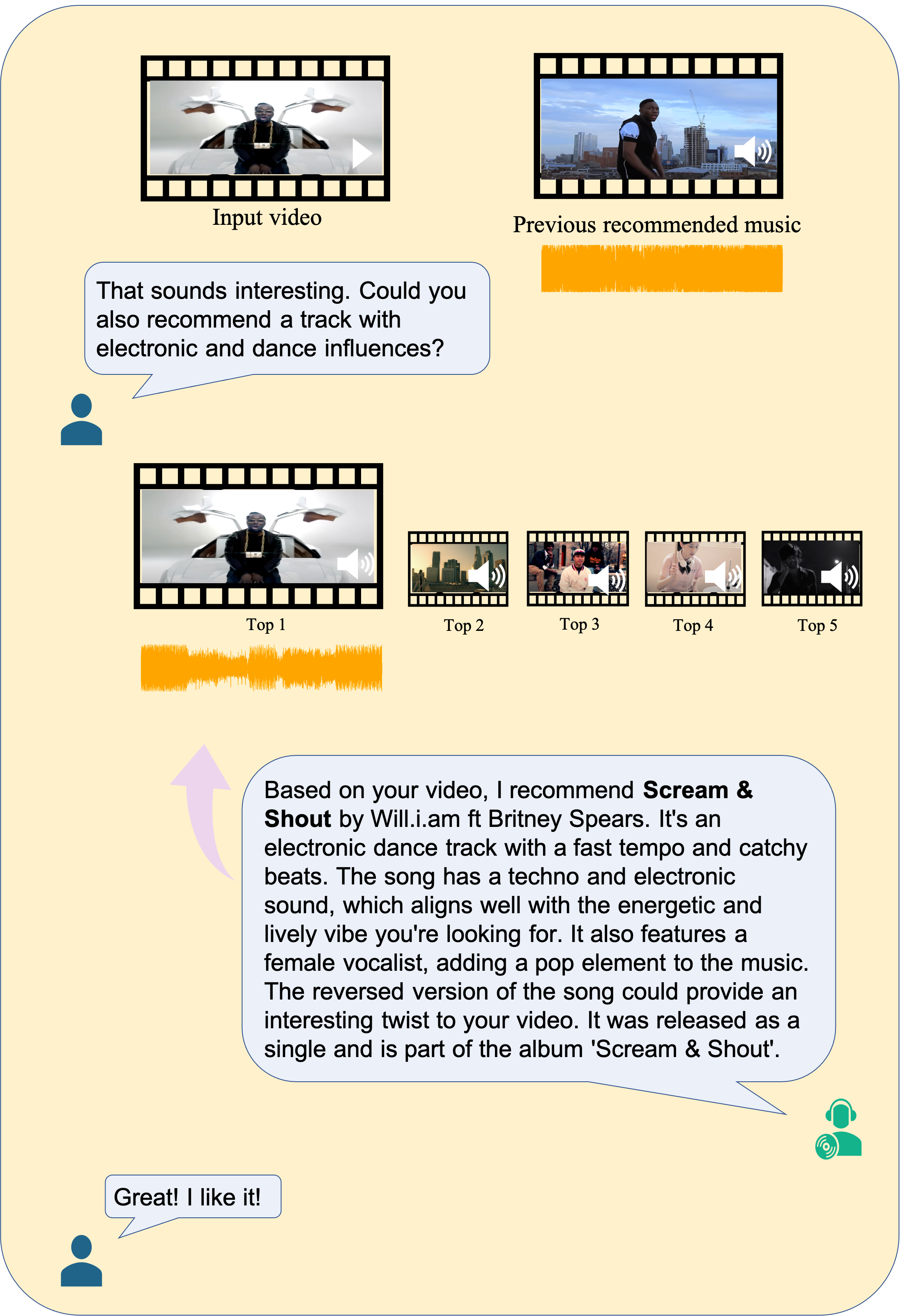}
    \caption{Example showing MuseChat's capability in improving recommendation results by learning user's preferences, contextual music and video content.}
    \label{fig:second_turn3}
\end{figure*}

We provide additional qualitative results in \Cref{fig:second_turn1,fig:second_turn2,fig:second_turn3}. 

%% file: main.bbl
\begin{thebibliography}{53}
\providecommand{\natexlab}[1]{#1}
\providecommand{\url}[1]{\texttt{#1}}
\expandafter\ifx\csname urlstyle\endcsname\relax
  \providecommand{\doi}[1]{doi: #1}\else
  \providecommand{\doi}{doi: \begingroup \urlstyle{rm}\Url}\fi

\bibitem[Abu-El-Haija et~al.(2016)Abu-El-Haija, Kothari, Lee, Natsev, Toderici, Varadarajan, and Vijayanarasimhan]{abu2016youtube}
Sami Abu-El-Haija, Nisarg Kothari, Joonseok Lee, Paul Natsev, George Toderici, Balakrishnan Varadarajan, and Sudheendra Vijayanarasimhan.
\newblock Youtube-8m: A large-scale video classification benchmark.
\newblock \emph{arXiv preprint arXiv:1609.08675}, 2016.

\bibitem[Bertin-Mahieux et~al.(2011)Bertin-Mahieux, Ellis, Whitman, and Lamere]{Bertin-Mahieux2011}
Thierry Bertin-Mahieux, Daniel~P.W. Ellis, Brian Whitman, and Paul Lamere.
\newblock The million song dataset.
\newblock In \emph{{Proceedings of the 12th International Conference on Music Information Retrieval ({ISMIR} 2011)}}, 2011.

\bibitem[Brown et~al.(2020)Brown, Mann, Ryder, Subbiah, Kaplan, Dhariwal, Neelakantan, Shyam, Sastry, Askell, et~al.]{brown2020language}
Tom Brown, Benjamin Mann, Nick Ryder, Melanie Subbiah, Jared~D Kaplan, Prafulla Dhariwal, Arvind Neelakantan, Pranav Shyam, Girish Sastry, Amanda Askell, et~al.
\newblock Language models are few-shot learners.
\newblock \emph{Advances in neural information processing systems}, 33:\penalty0 1877--1901, 2020.

\bibitem[Choi et~al.(2019)Choi, Lee, Park, and Nam]{choi2019zero}
Jeong Choi, Jongpil Lee, Jiyoung Park, and Juhan Nam.
\newblock Zero-shot learning for audio-based music classification and tagging.
\newblock \emph{arXiv preprint arXiv:1907.02670}, 2019.

\bibitem[Choi et~al.(2016)Choi, Fazekas, and Sandler]{choi2016automatic}
Keunwoo Choi, George Fazekas, and Mark Sandler.
\newblock Automatic tagging using deep convolutional neural networks.
\newblock \emph{arXiv preprint arXiv:1606.00298}, 2016.

\bibitem[Colombo et~al.(2022)Colombo, Clavel, and Piantanida]{colombo2022infolm}
Pierre Colombo, Chloe Clavel, and Pablo Piantanida.
\newblock Infolm: A new metric to evaluate summarization \& data2text generation, 2022.

\bibitem[Deng et~al.(2023)Deng, Zhang, Xu, Lei, Chua, and Lam]{deng2023unified}
Yang Deng, Wenxuan Zhang, Weiwen Xu, Wenqiang Lei, Tat-Seng Chua, and Wai Lam.
\newblock A unified multi-task learning framework for multi-goal conversational recommender systems.
\newblock \emph{ACM Transactions on Information Systems}, 41\penalty0 (3):\penalty0 1--25, 2023.

\bibitem[Dettmers et~al.(2023)Dettmers, Pagnoni, Holtzman, and Zettlemoyer]{dettmers2023qlora}
Tim Dettmers, Artidoro Pagnoni, Ari Holtzman, and Luke Zettlemoyer.
\newblock Qlora: Efficient finetuning of quantized llms, 2023.

\bibitem[Devlin et~al.(2018)Devlin, Chang, Lee, and Toutanova]{devlin2018bert}
Jacob Devlin, Ming-Wei Chang, Kenton Lee, and Kristina Toutanova.
\newblock Bert: Pre-training of deep bidirectional transformers for language understanding.
\newblock \emph{arXiv preprint arXiv:1810.04805}, 2018.

\bibitem[Doh et~al.(2023)Doh, Won, Choi, and Nam]{doh2023toward}
SeungHeon Doh, Minz Won, Keunwoo Choi, and Juhan Nam.
\newblock Toward universal text-to-music retrieval.
\newblock In \emph{ICASSP 2023-2023 IEEE International Conference on Acoustics, Speech and Signal Processing (ICASSP)}, pages 1--5. IEEE, 2023.

\bibitem[Friedman et~al.(2023)Friedman, Ahuja, Allen, Tan, Sidahmed, Long, Xie, Schubiner, Patel, Lara, et~al.]{friedman2023leveraging}
Luke Friedman, Sameer Ahuja, David Allen, Terry Tan, Hakim Sidahmed, Changbo Long, Jun Xie, Gabriel Schubiner, Ajay Patel, Harsh Lara, et~al.
\newblock Leveraging large language models in conversational recommender systems.
\newblock \emph{arXiv preprint arXiv:2305.07961}, 2023.

\bibitem[Gao et~al.(2021)Gao, Lei, He, de~Rijke, and Chua]{gao2021advances}
Chongming Gao, Wenqiang Lei, Xiangnan He, Maarten de Rijke, and Tat-Seng Chua.
\newblock Advances and challenges in conversational recommender systems: A survey.
\newblock \emph{AI Open}, 2:\penalty0 100--126, 2021.

\bibitem[Gao et~al.(2023{\natexlab{a}})Gao, Han, Zhang, Lin, Geng, Zhou, Zhang, Lu, He, Yue, Li, and Qiao]{gao2023llamaadapterv2}
Peng Gao, Jiaming Han, Renrui Zhang, Ziyi Lin, Shijie Geng, Aojun Zhou, Wei Zhang, Pan Lu, Conghui He, Xiangyu Yue, Hongsheng Li, and Yu Qiao.
\newblock Llama-adapter v2: Parameter-efficient visual instruction model.
\newblock \emph{arXiv preprint arXiv:2304.15010}, 2023{\natexlab{a}}.

\bibitem[Gao et~al.(2023{\natexlab{b}})Gao, Sheng, Xiang, Xiong, Wang, and Zhang]{gao2023chat}
Yunfan Gao, Tao Sheng, Youlin Xiang, Yun Xiong, Haofen Wang, and Jiawei Zhang.
\newblock Chat-rec: Towards interactive and explainable llms-augmented recommender system.
\newblock \emph{arXiv preprint arXiv:2303.14524}, 2023{\natexlab{b}}.

\bibitem[Gong et~al.(2021)Gong, Chung, and Glass]{gong2021ast}
Yuan Gong, Yu-An Chung, and James Glass.
\newblock Ast: Audio spectrogram transformer.
\newblock \emph{arXiv preprint arXiv:2104.01778}, 2021.

\bibitem[Guzhov et~al.(2022)Guzhov, Raue, Hees, and Dengel]{guzhov2022audioclip}
Andrey Guzhov, Federico Raue, J{\"o}rn Hees, and Andreas Dengel.
\newblock Audioclip: Extending clip to image, text and audio.
\newblock In \emph{ICASSP 2022-2022 IEEE International Conference on Acoustics, Speech and Signal Processing (ICASSP)}, pages 976--980. IEEE, 2022.

\bibitem[Güçlü et~al.(2016)Güçlü, Thielen, Hanke, and van Gerven]{güçlü2016brains}
Umut Güçlü, Jordy Thielen, Michael Hanke, and Marcel A.~J. van Gerven.
\newblock Brains on beats, 2016.

\bibitem[Hu et~al.(2021)Hu, Shen, Wallis, Allen-Zhu, Li, Wang, Wang, and Chen]{hu2021lora}
Edward~J. Hu, Yelong Shen, Phillip Wallis, Zeyuan Allen-Zhu, Yuanzhi Li, Shean Wang, Lu Wang, and Weizhu Chen.
\newblock Lora: Low-rank adaptation of large language models, 2021.

\bibitem[Hu et~al.(2023)Hu, Xiang, Qin, and Tan]{hu2023audio}
Tao Hu, Xuyu Xiang, Jiaohua Qin, and Yun Tan.
\newblock Audio--text retrieval based on contrastive learning and collaborative attention mechanism.
\newblock \emph{Multimedia Systems}, pages 1--14, 2023.

\bibitem[Huang et~al.(2022)Huang, Jansen, Lee, Ganti, Li, and Ellis]{huang2022mulan}
Qingqing Huang, Aren Jansen, Joonseok Lee, Ravi Ganti, Judith~Yue Li, and Daniel~PW Ellis.
\newblock Mulan: A joint embedding of music audio and natural language.
\newblock \emph{arXiv preprint arXiv:2208.12415}, 2022.

\bibitem[Jannach et~al.(2021)Jannach, Manzoor, Cai, and Chen]{jannach2021survey}
Dietmar Jannach, Ahtsham Manzoor, Wanling Cai, and Li Chen.
\newblock A survey on conversational recommender systems.
\newblock \emph{ACM Computing Surveys (CSUR)}, 54\penalty0 (5):\penalty0 1--36, 2021.

\bibitem[Lee et~al.(2017)Lee, Park, Kim, and Nam]{lee2017sample}
Jongpil Lee, Jiyoung Park, Keunhyoung~Luke Kim, and Juhan Nam.
\newblock Sample-level deep convolutional neural networks for music auto-tagging using raw waveforms.
\newblock \emph{arXiv preprint arXiv:1703.01789}, 2017.

\bibitem[Lee et~al.(2020)Lee, Bryan, Salamon, Jin, and Nam]{lee2020disentangled}
Jongpil Lee, Nicholas~J Bryan, Justin Salamon, Zeyu Jin, and Juhan Nam.
\newblock Disentangled multidimensional metric learning for music similarity.
\newblock In \emph{ICASSP 2020-2020 IEEE International Conference on Acoustics, Speech and Signal Processing (ICASSP)}, pages 6--10. IEEE, 2020.

\bibitem[Lin et~al.(2023)Lin, Sung, Lei, Bansal, and Bertasius]{lin2023vision}
Yan-Bo Lin, Yi-Lin Sung, Jie Lei, Mohit Bansal, and Gedas Bertasius.
\newblock Vision transformers are parameter-efficient audio-visual learners.
\newblock In \emph{Proceedings of the IEEE/CVF Conference on Computer Vision and Pattern Recognition}, pages 2299--2309, 2023.

\bibitem[Liu et~al.(2023)Liu, Dong, and Zhang]{liu2023tackling}
Xiulong Liu, Zhikang Dong, and Peng Zhang.
\newblock Tackling data bias in music-avqa: Crafting a balanced dataset for unbiased question-answering.
\newblock \emph{arXiv preprint arXiv:2310.06238}, 2023.

\bibitem[Liu et~al.(2021)Liu, Lin, Cao, Hu, Wei, Zhang, Lin, and Guo]{liu2021swin}
Ze Liu, Yutong Lin, Yue Cao, Han Hu, Yixuan Wei, Zheng Zhang, Stephen Lin, and Baining Guo.
\newblock Swin transformer: Hierarchical vision transformer using shifted windows.
\newblock In \emph{Proceedings of the IEEE/CVF international conference on computer vision}, pages 10012--10022, 2021.

\bibitem[Loshchilov and Hutter(2017)]{loshchilov2017decoupled}
Ilya Loshchilov and Frank Hutter.
\newblock Decoupled weight decay regularization.
\newblock \emph{arXiv preprint arXiv:1711.05101}, 2017.

\bibitem[Manco et~al.(2022{\natexlab{a}})Manco, Benetos, Quinton, and Fazekas]{manco2022contrastive}
Ilaria Manco, Emmanouil Benetos, Elio Quinton, and Gy{\"o}rgy Fazekas.
\newblock Contrastive audio-language learning for music.
\newblock \emph{arXiv preprint arXiv:2208.12208}, 2022{\natexlab{a}}.

\bibitem[Manco et~al.(2022{\natexlab{b}})Manco, Weck, Tovstogan, Won, and Bogdanov]{manco2022song}
Ilaria Manco, Benno Weck, Philip Tovstogan, Minz Won, and Dmitry Bogdanov.
\newblock Song describer: a platform for collecting textual descriptions of music recordings.
\newblock In \emph{Ismir 2022 Hybrid Conference}, 2022{\natexlab{b}}.

\bibitem[McKee et~al.(2023)McKee, Salamon, Sivic, and Russell]{mckee2023language}
Daniel McKee, Justin Salamon, Josef Sivic, and Bryan Russell.
\newblock Language-guided music recommendation for video via prompt analogies.
\newblock In \emph{Proceedings of the IEEE/CVF Conference on Computer Vision and Pattern Recognition}, pages 14784--14793, 2023.

\bibitem[Oord et~al.(2018)Oord, Li, and Vinyals]{oord2018representation}
Aaron van~den Oord, Yazhe Li, and Oriol Vinyals.
\newblock Representation learning with contrastive predictive coding.
\newblock \emph{arXiv preprint arXiv:1807.03748}, 2018.

\bibitem[Pons and Serra(2019)]{pons2019musicnn}
Jordi Pons and Xavier Serra.
\newblock musicnn: Pre-trained convolutional neural networks for music audio tagging.
\newblock \emph{arXiv preprint arXiv:1909.06654}, 2019.

\bibitem[Pr{\'e}tet et~al.(2021)Pr{\'e}tet, Richard, and Peeters]{pretet2021cross}
Laure Pr{\'e}tet, Gael Richard, and Geoffroy Peeters.
\newblock Cross-modal music-video recommendation: A study of design choices.
\newblock In \emph{2021 International Joint Conference on Neural Networks (IJCNN)}, pages 1--9. IEEE, 2021.

\bibitem[Radford et~al.(2018)Radford, Narasimhan, Salimans, Sutskever, et~al.]{radford2018improving}
Alec Radford, Karthik Narasimhan, Tim Salimans, Ilya Sutskever, et~al.
\newblock Improving language understanding by generative pre-training.
\newblock 2018.

\bibitem[Radford et~al.(2021)Radford, Kim, Hallacy, Ramesh, Goh, Agarwal, Sastry, Askell, Mishkin, Clark, et~al.]{radford2021learning}
Alec Radford, Jong~Wook Kim, Chris Hallacy, Aditya Ramesh, Gabriel Goh, Sandhini Agarwal, Girish Sastry, Amanda Askell, Pamela Mishkin, Jack Clark, et~al.
\newblock Learning transferable visual models from natural language supervision.
\newblock In \emph{International conference on machine learning}, pages 8748--8763. PMLR, 2021.

\bibitem[Sur{\'i}s et~al.(2022)Sur{\'i}s, Vondrick, Russell, and Salamon]{suris2022time}
D{\'i}dac Sur{\'i}s, Carl Vondrick, Bryan Russell, and Justin Salamon.
\newblock It’s time for artistic correspondence in music and video.
\newblock In \emph{Proceedings of the IEEE/CVF Conference on Computer Vision and Pattern Recognition}, pages 10564--10574, 2022.

\bibitem[Thoppilan et~al.(2022)Thoppilan, De~Freitas, Hall, Shazeer, Kulshreshtha, Cheng, Jin, Bos, Baker, Du, et~al.]{thoppilan2022lamda}
Romal Thoppilan, Daniel De~Freitas, Jamie Hall, Noam Shazeer, Apoorv Kulshreshtha, Heng-Tze Cheng, Alicia Jin, Taylor Bos, Leslie Baker, Yu Du, et~al.
\newblock Lamda: Language models for dialog applications.
\newblock \emph{arXiv preprint arXiv:2201.08239}, 2022.

\bibitem[Tian et~al.(2020)Tian, Krishnan, and Isola]{tian2020contrastive}
Yonglong Tian, Dilip Krishnan, and Phillip Isola.
\newblock Contrastive multiview coding.
\newblock In \emph{Computer Vision--ECCV 2020: 16th European Conference, Glasgow, UK, August 23--28, 2020, Proceedings, Part XI 16}, pages 776--794. Springer, 2020.

\bibitem[Touvron et~al.(2023)Touvron, Martin, Stone, Albert, Almahairi, Babaei, Bashlykov, Batra, Bhargava, Bhosale, et~al.]{touvron2023llama}
Hugo Touvron, Louis Martin, Kevin Stone, Peter Albert, Amjad Almahairi, Yasmine Babaei, Nikolay Bashlykov, Soumya Batra, Prajjwal Bhargava, Shruti Bhosale, et~al.
\newblock Llama 2: Open foundation and fine-tuned chat models.
\newblock \emph{arXiv preprint arXiv:2307.09288}, 2023.

\bibitem[Vaswani et~al.(2017)Vaswani, Shazeer, Parmar, Uszkoreit, Jones, Gomez, Kaiser, and Polosukhin]{vaswani2017attention}
Ashish Vaswani, Noam Shazeer, Niki Parmar, Jakob Uszkoreit, Llion Jones, Aidan~N Gomez, {\L}ukasz Kaiser, and Illia Polosukhin.
\newblock Attention is all you need.
\newblock \emph{Advances in neural information processing systems}, 30, 2017.

\bibitem[Wang et~al.(2023)Wang, Tang, Zhao, Wang, and Wen]{wang2023rethinking}
Xiaolei Wang, Xinyu Tang, Wayne~Xin Zhao, Jingyuan Wang, and Ji-Rong Wen.
\newblock Rethinking the evaluation for conversational recommendation in the era of large language models.
\newblock \emph{arXiv preprint arXiv:2305.13112}, 2023.

\bibitem[Won et~al.(2020)Won, Ferraro, Bogdanov, and Serra]{won2020evaluation}
Minz Won, Andres Ferraro, Dmitry Bogdanov, and Xavier Serra.
\newblock Evaluation of cnn-based automatic music tagging models.
\newblock \emph{arXiv preprint arXiv:2006.00751}, 2020.

\bibitem[Won et~al.(2021)Won, Choi, and Serra]{won2021semi}
Minz Won, Keunwoo Choi, and Xavier Serra.
\newblock Semi-supervised music tagging transformer.
\newblock \emph{arXiv preprint arXiv:2111.13457}, 2021.

\bibitem[Wu et~al.(2022)Wu, Seetharaman, Kumar, and Bello]{wu2022wav2clip}
Ho-Hsiang Wu, Prem Seetharaman, Kundan Kumar, and Juan~Pablo Bello.
\newblock Wav2clip: Learning robust audio representations from clip.
\newblock In \emph{ICASSP 2022-2022 IEEE International Conference on Acoustics, Speech and Signal Processing (ICASSP)}, pages 4563--4567. IEEE, 2022.

\bibitem[Yang et~al.(2021)Yang, Han, Li, Zuo, and Yu]{yang2021improving}
Bowen Yang, Cong Han, Yu Li, Lei Zuo, and Zhou Yu.
\newblock Improving conversational recommendation systems' quality with context-aware item meta information.
\newblock \emph{arXiv preprint arXiv:2112.08140}, 2021.

\bibitem[Yi et~al.(2021)Yi, Zhu, Xie, and Chen]{yi2021cross}
Jing Yi, Yaochen Zhu, Jiayi Xie, and Zhenzhong Chen.
\newblock Cross-modal variational auto-encoder for content-based micro-video background music recommendation.
\newblock \emph{IEEE Transactions on Multimedia}, 2021.

\bibitem[Zeng et~al.(2018)Zeng, Yu, and Oyama]{zeng2018audiovisual}
Donghuo Zeng, Yi Yu, and Keizo Oyama.
\newblock Audio-visual embedding for cross-modal music video retrieval through supervised deep cca.
\newblock In \emph{2018 IEEE International Symposium on Multimedia (ISM)}, pages 143--150. IEEE, 2018.

\bibitem[Zhang et~al.(2023{\natexlab{a}})Zhang, Li, and Bing]{zhang2023video}
Hang Zhang, Xin Li, and Lidong Bing.
\newblock Video-llama: An instruction-tuned audio-visual language model for video understanding.
\newblock \emph{arXiv preprint arXiv:2306.02858}, 2023{\natexlab{a}}.

\bibitem[Zhang et~al.(2023{\natexlab{b}})Zhang, Han, Liu, Gao, Zhou, Hu, Yan, Lu, Li, and Qiao]{zhang2023llamaadapter}
Renrui Zhang, Jiaming Han, Chris Liu, Peng Gao, Aojun Zhou, Xiangfei Hu, Shilin Yan, Pan Lu, Hongsheng Li, and Yu Qiao.
\newblock Llama-adapter: Efficient fine-tuning of language models with zero-init attention.
\newblock \emph{arXiv preprint arXiv:2303.16199}, 2023{\natexlab{b}}.

\bibitem[Zhang et~al.(2020)Zhang, Kishore, Wu, Weinberger, and Artzi]{zhang2020bertscore}
Tianyi Zhang, Varsha Kishore, Felix Wu, Kilian~Q. Weinberger, and Yoav Artzi.
\newblock Bertscore: Evaluating text generation with bert, 2020.

\bibitem[Zhao et~al.(2022)Zhao, Zhang, Zhu, Ma, and Zhang]{zhao2022s3t}
Hang Zhao, Chen Zhang, Bilei Zhu, Zejun Ma, and Kejun Zhang.
\newblock S3t: Self-supervised pre-training with swin transformer for music classification.
\newblock In \emph{ICASSP 2022-2022 IEEE International Conference on Acoustics, Speech and Signal Processing (ICASSP)}, pages 606--610. IEEE, 2022.

\bibitem[Zheng et~al.(2023)Zheng, Chiang, Sheng, Zhuang, Wu, Zhuang, Lin, Li, Li, Xing, Zhang, Gonzalez, and Stoica]{zheng2023judging}
Lianmin Zheng, Wei-Lin Chiang, Ying Sheng, Siyuan Zhuang, Zhanghao Wu, Yonghao Zhuang, Zi Lin, Zhuohan Li, Dacheng Li, Eric.~P Xing, Hao Zhang, Joseph~E. Gonzalez, and Ion Stoica.
\newblock Judging llm-as-a-judge with mt-bench and chatbot arena, 2023.

\bibitem[Zhu et~al.(2023)Zhu, Chen, Shen, Li, and Elhoseiny]{zhu2023minigpt}
Deyao Zhu, Jun Chen, Xiaoqian Shen, Xiang Li, and Mohamed Elhoseiny.
\newblock Minigpt-4: Enhancing vision-language understanding with advanced large language models.
\newblock \emph{arXiv preprint arXiv:2304.10592}, 2023.

\end{thebibliography}
